\documentclass[10pt,twocolumn,letterpaper]{article}

\usepackage[pagenumbers]{wacv} 
\usepackage[accsupp]{axessibility}  

\definecolor{wacvblue}{rgb}{0.21,0.49,0.74}
\usepackage[pagebackref,breaklinks,colorlinks,allcolors=wacvblue]{hyperref}
\usepackage{graphicx}
\usepackage{stfloats}
\usepackage[table,xcdraw]{xcolor}
\usepackage{colortbl}
\usepackage{booktabs}
\usepackage{tikz}
\usepackage{pgfplots}
\usepackage{multirow}
\usepackage{subcaption}
\usepackage{titletoc}
\usetikzlibrary{backgrounds,decorations.pathreplacing,positioning}
\usetikzlibrary{arrows.meta,positioning,calc}

\definecolor{SoftBlue}{HTML}{DAE8FC}
\definecolor{SoftYellow}{HTML}{FFF2CC}
\definecolor{SoftGreen}{HTML}{D5E8D4}

\definecolor{MyBlue}{HTML}{4176b6}
\definecolor{MyYellow}{HTML}{FFD966}
\definecolor{MyGreen}{HTML}{67AB9F}
\definecolor{SampleBlue}{HTML}{3A6BB5}
\definecolor{ThresholdRed}{HTML}{C70039}
\definecolor{EventGray}{HTML}{585858}

\def\wacvPaperID{1336} 
\def\confName{WACV}
\def\confYear{2026}

\title{VADER: Towards Causal Video Anomaly Understanding with Relation-Aware Large Language Models}

\author{
    Ying Cheng\textsuperscript{1}, 
    Yu-Ho Lin\textsuperscript{1}, 
    Min-Hung Chen\textsuperscript{2}, 
    Fu-En Yang\textsuperscript{2},
    Shang-Hong Lai\textsuperscript{1}\thanks{Corresponding author}
    \\ 
    \vspace{0.5cm}
    \textsuperscript{1}National Tsing Hua University \quad 
    \textsuperscript{2}NVIDIA
}

\begin{document}
\maketitle
\begin{abstract}
Video anomaly understanding (VAU) aims to provide detailed interpretation and semantic comprehension of anomalous events within videos, addressing limitations of traditional methods that focus solely on detecting and localizing anomalies. However, existing approaches often neglect the deeper causal relationships and interactions between objects, which are critical for understanding anomalous behaviors. In this paper, we propose \textbf{VADER}, an LLM-driven framework for \textbf{V}ideo \textbf{A}nomaly un\textbf{DE}rstanding, which integrates keyframe object \textbf{R}elation features with visual cues to enhance anomaly comprehension from video. 
Specifically, VADER first applies an Anomaly Scorer to assign per-frame anomaly scores, followed by a Context-AwarE Sampling (CAES) strategy to capture the causal context of each anomalous event. A Relation Feature Extractor and a COntrastive Relation Encoder (CORE) jointly model dynamic object interactions, producing compact relational representations for downstream reasoning. These visual and relational cues are integrated with LLMs to generate detailed, causally grounded descriptions and support robust anomaly-related question answering. 
Experiments on multiple real-world VAU benchmarks demonstrate that VADER achieves strong results across anomaly description, explanation, and causal reasoning tasks, advancing the frontier of explainable video anomaly analysis. 
Project page is available at \url{https://vader-vau.github.io/}.
\end{abstract}    
\section{Introduction}
\label{sec:intro}

\begin{figure}[ht!]
    \centering
    \includegraphics[width=0.5\textwidth]{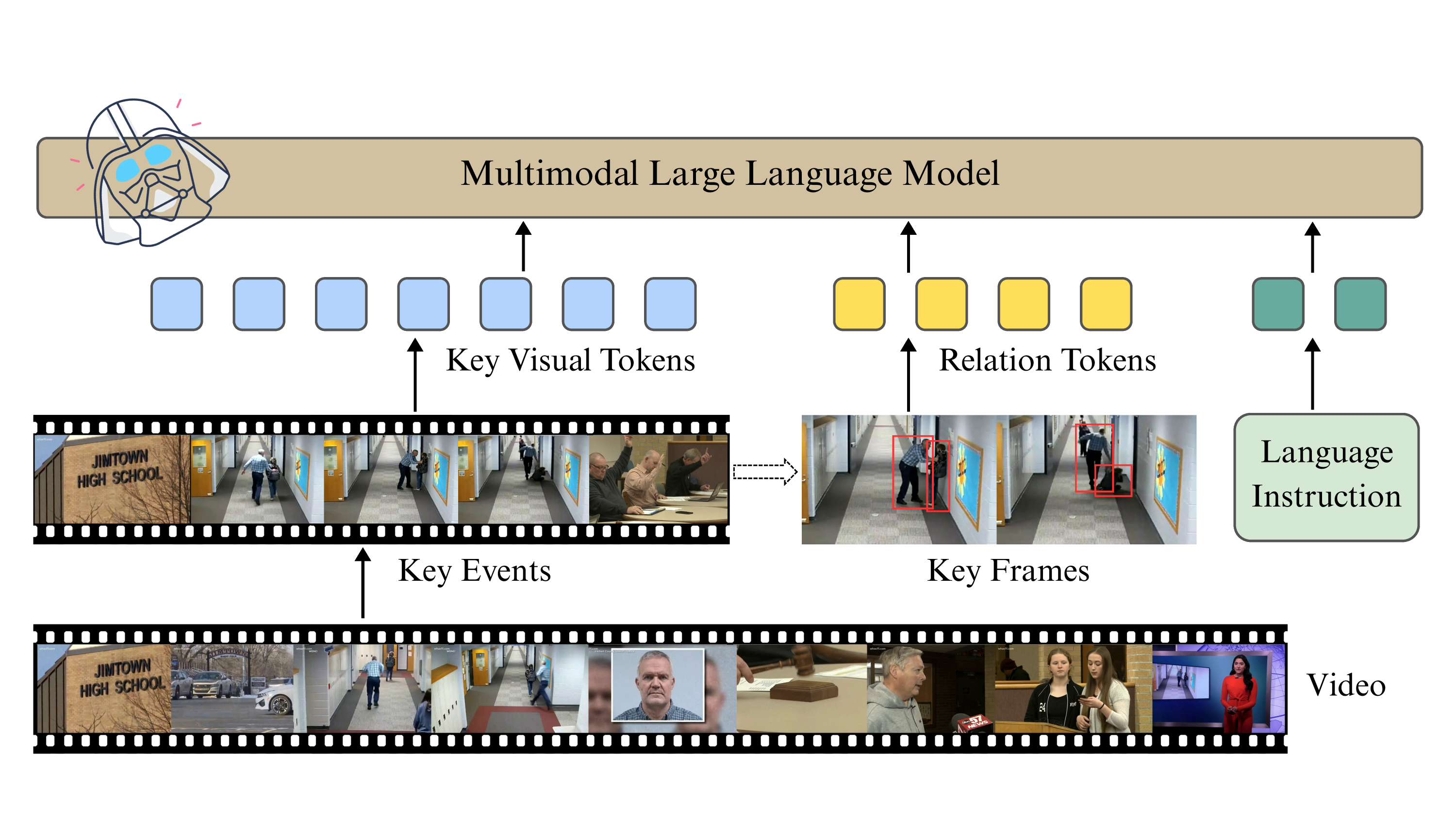}
    \caption{\textbf{Concept of VADER.} VADER enables detailed anomaly understanding by extracting key visual and relational cues from selected key frames.}
    \label{fig:teaser}
\end{figure}

Video anomaly detection (VAD) plays a critical role in surveillance, traffic monitoring, and public safety. Conventional methods primarily focus on temporal localization and classification \cite{sultani2018real, park2020learning, liu2023diversity, wu2022self, zhao2017spatio,wang2021robust, liu2021hybrid, georgescu2021anomaly}, but often lack semantic interpretability, limiting their utility in decision-making. To address this, recent research has shifted from video anomaly detection to video anomaly understanding (VAU)~\cite{yang2024anomalyruler, atang2024hawk, lv2024video, zhang2025holmes, wang2024visiongpt}, emphasizing detailed descriptions, causal reasoning, and context-aware explanations.

Recent advances in Large Language Models (LLMs) \cite{li2023blip, alayrac2022flamingo, hu2024minicpm, Maaz2023VideoChatGPT, ye2023mplug, liu2025nvila, liu2024oryx} have enabled natural language understanding of video content, paving the way for explainable VAU. Several works have leveraged LLMs for open-world VAU, including causation-focused pipeline~\cite{CUVA}, interaction-aware framework~\cite{atang2024hawk}, and temporal segment selection~\cite{zhang2025holmes}. However, these approaches often neglect deeper causal relationships and dynamic object interactions, which are critical for understanding unusual behaviors. See Supp. Material Sec.~\ref{sec:supp_motivation}.

In this work, we propose \textbf{VADER}, an LLM-driven framework for \textbf{V}ideo \textbf{A}nomaly un\textbf{DE}rstanding, integrating keyframe-level object \textbf{R}elation features and visual cues to enhance anomaly comprehension in video, as illustrated in Figure~\ref{fig:teaser}. 
VADER first applies an Anomaly Scorer to assign per-frame anomaly scores, followed by a Context-AwarE Sampling (\textbf{CAES}) strategy that adaptively selects keyframes to capture the full causal context of each event. 
For each sampled keyframe, a Relation Feature Extractor provides rich object and relational representations, which are further aggregated by a COntrastive Relation Encoder (\textbf{CORE}) into compact relational tokens that encode salient and temporally resolved interaction dynamics. 
By integrating both visual and relational cues into the LLM, VADER enables detailed, causally consistent narrative generation and robust anomaly-related question answering.

We extensively evaluate VADER on three recent and challenging VAU benchmarks, covering tasks such as anomaly description, question answering, and causal reasoning. Our method outperforms strong baselines and recent state-of-the-art systems across multiple metrics. Ablation analyses further highlight the impact of each major component.

In summary, our main contributions are as follows:
\begin{itemize}
\item We introduce VADER, an integrated framework that combines context-aware event sampling with dynamic relational modeling to support causally grounded video anomaly understanding with LLMs.
\item We propose a Context-AwarE Sampling strategy (CAES) based on anomaly scores and temporal gradients, enabling story-driven keyframe selection.
\item We develop a weakly supervised COntrastive Relation Encoder (CORE) that produces dynamic tokens for evolving object interactions, enhancing both interpretability and reasoning depth.
\item We validate VADER on multiple challenging benchmarks, achieving state-of-the-art or highly competitive performance in video anomaly description, explanation, and reasoning tasks.
\end{itemize}
\vspace{-0.07in}
\section{Related Work}
\label{sec:related_work}

\subsection{Video Anomaly Detection and Understanding}

Video Anomaly Detection (VAD) aims to identify events that deviate from typical patterns in videos. 
Early approaches~\cite{mehran2009abnormal, li2013anomaly, saligrama2012video, zaharescu2010anomalous} relied on handcrafted features and traditional algorithms to capture motion or appearance anomalies. 
With the advent of deep learning, CNNs and RNNs have been widely adopted to model spatiotemporal dynamics~\cite{liu2018future, chong2017abnormal, tran2015learning, zhao2017spatio} and enhance detection performance. 
Recent methods~\cite{hasan2015context, zhou2023dual, zhou2024batchnorm, nayak2021comprehensive, lv2023unbiased, zhou2023dual} often utilize reconstruction-based or weakly supervised approaches to address these tasks, aiming to learn normal patterns and detect deviations as anomalies. 
To facilitate progress in this field, several datasets \cite{acsintoae2022ubnormal, chan2008modeling, liu2018future, lu2013abnormal, sultani2018real, wang2010anomaly, yao2022dota} have been introduced, covering diverse video content and annotation granularity.
Beyond VAD, Video Anomaly Understanding (VAU) involves deeper analysis, such as describing anomaly contents or investigating their contexts. 
With recent advancements in text generation and multimodal learning, VAU research has seen significant progress. Several benchmarks such as~\cite{CUVA, atang2024hawk, yuan2023surveillance, zhang2025holmes} have been developed. Building on these resources, VAU emerged as a new task, and traditional VAD was further extended with MLLM-based approaches; we review these in the following section.

\subsection{Video Anomaly Analysis with MLLMs}

Recent advances in Multimodal Large Language Models (MLLMs)~\cite{alayrac2022flamingo, li2023blip, liu2023visual, zhu2023minigpt} have greatly expanded the scope of video anomaly analysis. 
By jointly modeling visual content and natural language, MLLMs enable capabilities such as open-vocabulary anomaly detection and reasoning~\cite{cao2025personalizing, xu2024large}.
Video-capable MLLMs~\cite{Maaz2023VideoChatGPT, liu2025nvila, hu2024minicpm, li2025otter, liu2024oryx} further enhance these abilities by reasoning over video data. Related works on VAD include LAVAD~\cite{zanella2024harnessing} and SUVAD~\cite{gao2025suvad} for training-free anomaly detection; AnomalyRuler~\cite{yang2024anomalyruler} with rule-based reasoning; VAD-LLaMA~\cite{lv2024video} and VERA~\cite{ye2025vera} as an explainable detector.
Works on VAU include Holmes-VAU~\cite{zhang2025holmes} with an anomaly-focused sampling; CUVA~\cite{CUVA} with causation-focused reasoning; HAWK~\cite{atang2024hawk} with interaction-aware modeling; and AssistPDA~\cite{yang2025assistpda} for real-time analysis.
Despite these advances, existing approaches still overlook the deeper causal relationships and interactions between objects, which are critical for understanding anomalous behaviors.

\subsection{Modeling Relationships with Scene Graphs}

Understanding object relationships is vital for high-level video analysis. Many anomalies arise from unusual interactions or collective behaviors, such as a person running against a crowd, which cannot be fully captured by appearance features alone.
Scene graphs provide a structured representation of these interactions by modeling objects as nodes and relationships as edges, enabling models to move beyond low-level features toward relational understanding. 
EGTR~\cite{im2024egtr} extracts relational information from DETR~\cite{carion2020end} for efficient graph generation, while DecoAD~\cite{chen2025unveiling} integrates scene and action features via knowledge graphs to better detect human-centric anomalies. Lohner et al.~\cite{lohner2024enhancing} further showed that adding encoded scene graph features improves detection performance, highlighting the value of explicit relational modeling.
Building on these advances, we leverage~\cite{im2024egtr} to extract object-level relations and integrate them into our framework, enabling more interpretable and comprehensive anomaly understanding.

\begin{figure*}[ht!]
    \centering
    \includegraphics[width=\textwidth]{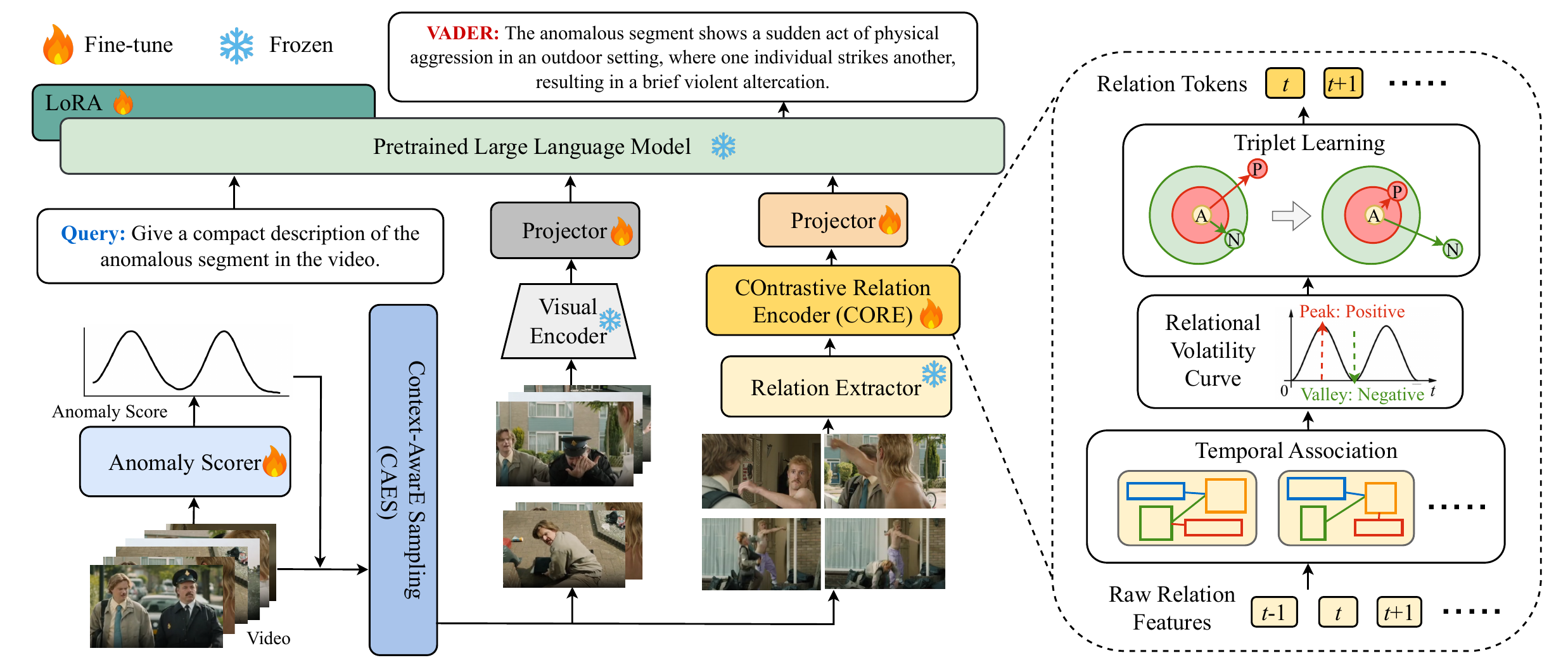}
    \caption{\textbf{Overview of VADER framework.}  Given an input video, the Anomaly Scorer and Context-AwarE Sampling (CAES) identify keyframes for narrative-driven anomaly analysis. Visual and relational features are extracted and encoded, with dynamic relational patterns distilled by the COntrastive Relation Encoder (CORE). All cues are fused by a pretrained LLM for comprehensive video anomaly understanding. The right panel illustrates the relational branch, including temporal association, volatility mining, and contrastive token learning.}
    \label{fig:workflow}
\end{figure*}

\section{Method}
\label{sec:method}

In this section, we present VADER, a novel framework for deep, causal understanding of anomalous events in complex videos. Unlike traditional anomaly detection approaches, VADER is designed to interpret and explain both the occurrence and underlying causes of anomalous events. Our framework consists of two main components: (1) keyframe detection and selection (Sec \ref{subsec:Keyframe}) for narrative-driven context sampling, and (2) relation extraction and integration (Sec \ref{subsec:Relation}) for fine-grained object interaction modeling. An overview of our pipeline is shown in Figure~\ref{fig:workflow}.

\subsection{Keyframe Detection and Selection}
\label{subsec:Keyframe}

To focus computational resources on the most informative video content, VADER first assigns an anomaly score to each frame using an Anomaly Scorer and then applies a keyframe selection strategy to filter and refine the candidate frames for downstream analysis.

\subsubsection{Anomaly Scoring}
To build a robust and widely applicable system, we leverage an Anomaly Scorer inspired by the CLIP-based framework \cite{zanella2024delving}. Rather than training a specialized model for each target dataset \cite{zhou2023dual,zhou2024batchnorm}, our scorer is pre-trained once on a large, aggregated dataset encompassing diverse video scenarios. 

Following~\cite{zanella2024delving}, we compute a normality prototype \( \mathbf{m} \) by averaging features extracted by the CLIP~\cite{radford2021learning} image encoder \( E_I \) on all normal training video frames \( \mathcal{I}_\text{norm} \) :
\[
\mathbf{m} = \frac{1}{|\mathcal{I}_\text{norm}|} \sum_{\mathbf{x} \in \mathcal{I}_\text{norm}} E_I(\mathbf{x})
, \tag{1}\
\]

Both visual and text features are then re-centered by subtracting \( \mathbf{m} \), aligning the origin with normal behavior, where distances indicate abnormality and directions encode semantics. The re-centered frame features are then projected onto class-specific semantic directions $ d_c $ and passed through a softmax to produce a conditional class distribution \( p_{c|A}(I_i) \), representing the probability of each anomaly class $ c $ given that an anomaly occurs.

The per-frame anomaly probability \( p_A(I_i) \), estimated by a temporal module that captures short- and long-term dependencies across frame sequences, is combined with the conditional class distribution \( p_{c|A}(I_i) \) to form a joint probability. The frame-level anomaly score is computed as: 
\[
S_i = \max_{c} \big(p_A(I_i) \cdot p_{c|A}(I_i)\big)
, \tag{2}\
\]

The resulting anomaly scores \( S \) are used to guide keyframe selection, while the predicted anomaly classes provide semantic cues for downstream reasoning by LLMs.

\subsubsection{Context-AwarE Sampling (CAES)}
To capture the narrative structure of anomalous events, we introduce CAES, a Context-AwarE keyframe Sampling strategy that preserves both causal context and event diversity. Formally, \[
K = \text{CAES}(S)
, \tag{3}\
\]
where \( K \) is the set of selected keyframes and \( S \) is the anomaly score sequence. 

\begin{figure}[t]
\centering
\resizebox{\columnwidth}{!}{
\begin{tikzpicture}[yscale=1.12]
    \definecolor{MyBlue}{HTML}{4176b6}
    \definecolor{MyYellow}{HTML}{FFD966}
    \definecolor{MyGreen}{HTML}{67AB9F}
    \definecolor{SampleBlue}{HTML}{3A6BB5}
    \definecolor{ThresholdRed}{HTML}{C70039}

    \def\xPreStart{0.7}
    \def\xRise{2.3}      
    \def\xPreEnd{4.5}
    \def\xOnEnd{7.1}
    \def\xCalm{8.2}      
    \def\xPostEnd{9.1}
    \def\xFlipPoint{9.8} 

    \fill[MyGreen!32, rounded corners=2pt]   ({\xFlipPoint-\xPreEnd},0) rectangle ({\xFlipPoint-\xRise},2.2); 
    \fill[MyYellow!60, rounded corners=2pt] ({\xFlipPoint-\xOnEnd},0) rectangle ({\xFlipPoint-\xPreEnd},3.2); 
    \fill[MyBlue!32, rounded corners=2pt]  ({\xFlipPoint-\xCalm},0) rectangle ({\xFlipPoint-\xOnEnd},1.8);  

    \draw[thick] (\xPreStart-0.1,0) -- (\xPostEnd+0.2,0);
    \draw[thick] (\xPreStart,0) -- (\xPreStart,3.5);
    \node[anchor=south west, font=\small] at (\xPreStart-0.08,3.45) {Anomaly Score};
    \node[anchor=west, font=\small] at (\xPostEnd+0.17,0) {Frame};

    \draw[thick,black, smooth, samples=100, domain=\xPreStart:\xPostEnd] 
        plot ({\xFlipPoint-\x}, {0.42 + 0.15*(\x-0.7)^2/8.4 + 1.08*exp(-(\x-6.1)^2/1.14) - 0.83*exp(-(\x-8.3)^2/0.33)});

    \foreach \i in {1,2} {
        \pgfmathsetmacro{\x}{\xRise+(\xPreEnd-\xRise)*\i/3}
        \pgfmathsetmacro{\y}{0.42 + 0.15*(\x-0.7)^2/8.4 + 1.08*exp(-(\x-6.1)^2/1.14) - 0.83*exp(-(\x-8.3)^2/0.33)}
        \fill[SampleBlue] ({\xFlipPoint-\x},\y) circle (2.3pt);
    }
    \foreach \i in {1,2,3,4} {
        \pgfmathsetmacro{\x}{\xPreEnd+(\xOnEnd-\xPreEnd)*\i/5}
        \pgfmathsetmacro{\y}{0.42 + 0.15*(\x-0.7)^2/8.4 + 1.08*exp(-(\x-6.1)^2/1.14) - 0.83*exp(-(\x-8.3)^2/0.33)}
        \fill[SampleBlue] ({\xFlipPoint-\x},\y) circle (2.3pt);
    }
    \foreach \i in {1,2} {
        \pgfmathsetmacro{\x}{\xOnEnd+(\xCalm-\xOnEnd)*\i/3}
        \pgfmathsetmacro{\y}{0.42 + 0.15*(\x-0.7)^2/8.4 + 1.08*exp(-(\x-6.1)^2/1.14) - 0.83*exp(-(\x-8.3)^2/0.33)}
        \fill[SampleBlue] ({\xFlipPoint-\x},\y) circle (2.3pt);
    }

    \pgfmathsetmacro{\yrise}{0.42 + 0.15*(\xRise-0.7)^2/8.4 + 1.08*exp(-(\xRise-6.1)^2/1.14) - 0.83*exp(-(\xRise-8.3)^2/0.33)}
    \pgfmathsetmacro{\ycalm}{0.42 + 0.15*(\xCalm-0.7)^2/8.4 + 1.08*exp(-(\xCalm-6.1)^2/1.14) - 0.83*exp(-(\xCalm-8.3)^2/0.33)}

    \fill[ThresholdRed] ({\xFlipPoint-\xCalm},\ycalm) circle (1.2pt);
    \node[ThresholdRed, font=\tiny, anchor=east] at ({\xFlipPoint-\xCalm-0.05}, \ycalm+0.01) {Rise threshold};
    
    \fill[ThresholdRed] ({\xFlipPoint-\xRise},\yrise) circle (1.2pt);
    \node[ThresholdRed, font=\tiny, anchor=east] at ({\xFlipPoint-1.4}, \yrise-0.13) {Calm threshold};

    \node[black, font=\bfseries, opacity=0.87, scale=0.94] at ({\xFlipPoint-3.4},2.00) {Post-Event};
    \node[black, font=\bfseries, opacity=0.93, scale=0.96] at ({\xFlipPoint-5.8},3.00) {On-Event};
    \node[black, font=\bfseries, opacity=0.91, scale=0.94] at ({\xFlipPoint-7.6},1.60) {Pre-Event};
    
    \node[MyGreen!70!black, font=\scriptsize] at ({\xFlipPoint-3}, -0.21) {Falling Slope (post-event)};
    \node[black, font=\scriptsize] at ({\xFlipPoint-5.6}, -0.21) {Peak (on-event)};
    \node[MyBlue!70!black, font=\scriptsize] at ({\xFlipPoint-8.0}, -0.21) {Rising Slope (pre-event)};
    
    \draw[dashed, line width=0.7pt] ({\xFlipPoint-\xPreEnd},0) -- ({\xFlipPoint-\xPreEnd},2.2) -- ({\xFlipPoint-\xRise},2.2) -- ({\xFlipPoint-\xRise},0);
    \draw[dashed, line width=0.7pt] ({\xFlipPoint-\xOnEnd},0) -- ({\xFlipPoint-\xOnEnd},3.2) -- ({\xFlipPoint-\xPreEnd},3.2) -- ({\xFlipPoint-\xPreEnd},0);
    \draw[dashed, line width=0.7pt] ({\xFlipPoint-\xCalm},0) -- ({\xFlipPoint-\xCalm},1.8) -- ({\xFlipPoint-\xOnEnd},1.8) -- ({\xFlipPoint-\xOnEnd},0);
\end{tikzpicture}
}
\vspace{-2mm}
\caption{\textbf{Illustration of CAES keyframe selection strategy.} The anomaly score curve is segmented into pre-event (blue), on-event (yellow), and post-event (green) intervals. Blue dots are sampled keyframes, and red dots indicate rise and calm thresholds.}
\label{fig:caes_sampling}
\end{figure}
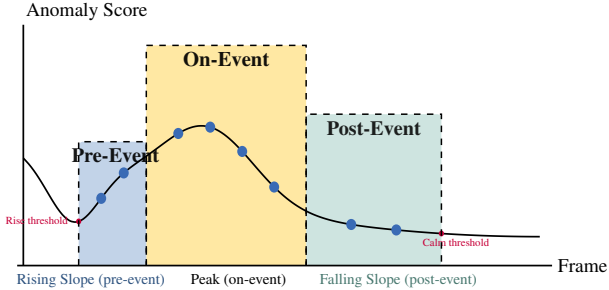

As illustrated in Figure~\ref{fig:caes_sampling}, CAES first detects all anomalous intervals using a percentile-based adaptive threshold for each video, then automatically expands each event into pre-event and post-event context segments by applying rise and calm thresholds to the slope of anomaly scores. Keyframes are then uniformly sampled from pre-event, on-event, and post-event segments to cover the causal lead-up, climax, and aftermath of each anomaly, ensuring a comprehensive yet compact representation.

If multiple anomalies are detected or the total exceeds the frame budget (e.g., 64), frames with the highest anomaly scores are prioritized. Any remaining slots are filled by sampling from background segments to ensure temporal coherence. This approach ensures a comprehensive yet compact representation for downstream narrative reasoning.

\subsection{Relation Extraction and Integration}
\label{subsec:Relation}
While CAES identifies what and when to analyze, the core of our VADER framework lies in its ability to understand how the relationships between objects dynamically evolve. Moving beyond static scene description, VADER first extracts rich object and relational features from each selected frame, then automatically discovers temporal patterns of interaction using only video-level labels. Finally, it distills these patterns into compact dynamic representations and integrates them into the main LLM for deep, causally-grounded reasoning about complex events. 

\subsubsection{Relational Feature Extraction}
For each keyframe selected by CAES, we utilize a pre-trained DETR-based scene graph generation model \cite{im2024egtr} as a feature extractor to extract appearance embeddings for detected objects and a relational tensor. This tensor contains the intermediate features computed for each object pair before the relationship classification head, serving as a comprehensive snapshot of the scene's relational dynamics. 

To maintain consistent object identities across frames, we perform object association using a combination of IoU overlap and cosine similarity of appearance embeddings. 
To capture temporal changes in these interactions, we compute a relational volatility curve for each video by tracking objects and, for each pair of adjacent frames, taking the maximum $L_2$ norm of the difference between relational features for all co-tracked object pairs:
\[
\text{Volatility}(t) = \max_{(i, j)} \left\| \mathbf{r}_{ij}(t) - \mathbf{r}_{ij}(t-1) \right\|_2
, \tag{4}\
\]
This sequence of volatility scores, together with the associated before/after relational vectors, is cached for efficient downstream processing.

\subsubsection{COntrative Relation Encoder (CORE)}
Based on the relational volatility curves, we automatically mine and encode salient patterns of object interaction change in a weakly-supervised manner. 
To ensure robust peak detection, each volatility curve is first smoothed with a Gaussian filter. 
For abnormal videos, relational-change pairs, i.e., $[\mathbf{r}{ij}(t-1); \mathbf{r}{ij}(t)]$, corresponding to the top k\% of these smoothed peaks are designated as positive samples, while pairs from volatility valleys and from normal videos serve as negatives.

To distill these transitions into compact representations, we train a lightweight COntrastive Relation Encoder (CORE) that maps each relational-change pair to compact relation tokens. To train CORE, we employ a triplet margin loss, formulated as
\[
\mathcal{L}_{\text{triplet}} = \max \left(0, d(f_a, f_p) - d(f_a, f_n) + \alpha \right)
, \tag{5}\
\]
where $f_a$, $f_p$, and $f_n$ denote the anchor, positive, and negative samples encoded by CORE, $d(\cdot, \cdot)$ is the L2 distance, and $\alpha$ is the margin. We further adopt semi-hard negative mining~\cite{schroff2015facenet} to improve discriminative power.

\subsection{LLM Integration and Fine-tuning}
For each video, keyframes selected by CAES are encoded into visual tokens and passed through the Relation Feature Extractor to obtain raw relational tensors, from which relation tokens are generated by CORE. Task instructions are converted into text tokens. All tokens are concatenated and jointly fed into the Multimodal LLM, enabling the model to reason over both visual content and the evolving object relationships that underlie complex events.

During training, we fine-tune only the multimodal projectors and LoRA~\cite{hu2022lora} adapters, keeping the backbone frozen. This setup allows VADER to align both static and dynamic video cues with causal, semantically rich natural language output. As a result, the model can generate detailed, context-aware anomaly descriptions and perform robust anomaly-related reasoning.

\vspace{0.3em}
\noindent\textbf{Implementation Details.} 
Including hyperparameters, threshold values, and training configurations, are provided in the Suppl. Material Sec.~\ref{sec:supp_implementaiton_details}.
\section{Experiments}
In this section, we present the experimental results of applying the proposed VADER model to three latest comprehensive benchmark datasets commonly used for evaluating VAU methods. We also compare the experimental results with the SOTA methods and discuss some ablation studies on the proposed method.

\begin{table}[h]
\centering
\renewcommand{\arraystretch}{1.3}
\resizebox{\columnwidth}{!}{
\begin{tabular}{lccccccc}
\toprule
\textbf{Benchmark} & \cellcolor{SoftBlue}BLEU & \cellcolor{SoftBlue}ROUGE & \cellcolor{SoftBlue}CIDEr & \cellcolor{SoftBlue}METEOR & \cellcolor{SoftYellow}MoverScore & \cellcolor{SoftGreen}GPT Score & \cellcolor{SoftGreen}MMEval \\
\cmidrule(lr){2-5} \cmidrule(lr){6-6} \cmidrule(lr){7-8}
& \multicolumn{4}{c}{\cellcolor{SoftBlue}\textit{Lexical-level}} & \cellcolor{SoftYellow}\textit{Semantic-level} & \multicolumn{2}{c}{\cellcolor{SoftGreen}\textit{Judge-based}} \\
\midrule
HIVAU-70k & \checkmark & \checkmark & \checkmark & \checkmark & & & \\
HAWK      & \checkmark &           &           &           & & \checkmark & \\
CUVA      & \checkmark & \checkmark &           &           & \checkmark & & \checkmark \\
\bottomrule
\end{tabular}
}
\caption{Evaluation metrics for each benchmark. Metric types are grouped into lexical-level (blue), semantic-level (yellow), and judge-based (green).}
\label{tab:metrics_summary}
\end{table}

\renewcommand{\arraystretch}{1.1}
\begin{table}
    \centering
    \resizebox{\columnwidth}{!}{ 
    \begin{tabular}{l l ccc}
        \toprule
        \textbf{Method} & \textbf{Metric} & \textbf{Description} & \textbf{Causes} & \textbf{Effect} \\
        \midrule
        mPLUG-owl \cite{ye2023mplug} & BLEU & 0.55 & 0.65 & 0.47 \\
        & ROUGE & 12.58 & 13.54 & 8.83 \\
        & BLEURT & 40.66 & 43.28 & 37.95 \\
        & MoverScore & 51.97 & 52.71 & 50.06 \\
        & UniEval & 67.46 & 62.29 & \underline{59.07} \\
        & MMEval & 73.42 & 17.15 & 44.31 \\
        \midrule
        Video-LLAMA \cite{damonlpsg2023videollama} & BLEU & 0.60 & 0.53 & 0.35 \\
        & ROUGE & 13.15 & 12.36 & 8.02 \\
        & BLEURT & 40.55 & 43.02 & 39.68 \\
        & MoverScore & 51.32 & 51.25 & 49.48 \\
        & UniEval & 52.28 & 47.29 & 43.03 \\
        & MMEval & 65.65 & 16.24 & 32.84 \\
        \midrule
        PandaGPT \cite{su2023pandagpt} & BLEU & 0.66 & 0.51 & 0.30 \\
        & ROUGE & 13.33 & 14.09 & 8.79 \\
        & BLEURT & 38.23 & 43.95 & \underline{39.95} \\
        & MoverScore & 51.73 & 51.54 & 49.62 \\
        & UniEval & 57.05 & 54.88 & 50.84 \\
        & MMEval & 74.19 & 22.47 & \textbf{69.45} \\
        \midrule
        Otter \cite{li2025otter} & BLEU & \underline{1.07} & \underline{1.09} & \textbf{1.11} \\
        & ROUGE & \textbf{15.19} & \underline{15.87} & \underline{11.40} \\
        & BLEURT & 29.92 & 32.52 & 28.94 \\
        & MoverScore & \textbf{53.54} & \textbf{54.25} & \underline{51.91} \\
        & UniEval & 45.14 & 49.05 & 47.51 \\
        & MMEval & 76.30 & 3.53 & 39.21 \\
        \midrule
        Video-ChatGPT \cite{Maaz2023VideoChatGPT} & BLEU & 0.30 & 0.29 & 0.41 \\
        & ROUGE & 9.75 & 9.08 & 8.23 \\
        & BLEURT & \underline{46.83} & \textbf{49.52} & 37.24 \\
        & MoverScore & 50.73 & 50.70 & 49.83 \\
        & UniEval & \underline{70.82} & \underline{70.77} & 54.35 \\
        & MMEval & 78.55 & 44.57 & 46.08 \\
        \midrule
        CUVA$^{*}$ \cite{CUVA} & BLEU & 0.55 & 0.51 & 0.38 \\
        & ROUGE & \underline{14.35} & 9.08 & 8.23 \\
        & BLEURT & \textbf{47.10} & \underline{48.13} & \textbf{48.28} \\
        & MoverScore & 52.25 & 52.28 & 49.95 \\
        & UniEval & 68.18 & 63.41 & 51.87 \\
        & MMEval & \textbf{79.65} & \underline{58.92} & 50.64 \\
        \midrule
        \rowcolor{SoftBlue}\textbf{VADER$^{\dag}$ (Ours)} & BLEU & \textbf{1.09} & \textbf{1.19} & \textbf{1.11} \\
        \rowcolor{SoftBlue}& ROUGE & 12.85 & \textbf{16.84} & \textbf{17.38} \\
        \rowcolor{SoftBlue}& BLEURT & 32.34 & 37.48 & 37.60 \\
        \rowcolor{SoftBlue}& MoverScore & \underline{52.65} & \underline{52.78} & \textbf{53.21} \\
        \rowcolor{SoftBlue}& UniEval & \textbf{85.00} & \textbf{79.37} & \textbf{82.06} \\
        \rowcolor{SoftBlue}& MMEval & \underline{78.89} & \textbf{66.30} & \underline{63.26} \\
        \bottomrule
    \end{tabular}}
    \caption{\textbf{Evaluation on the CUVA benchmark.} VADER shows competitive results across anomaly description and causation. \textbf{Bold} indicate the best performance, and \textit{underlined} indicate the second best. Models marked with $^{\dag}$ are finetuned on CUVA; those marked with $^{*}$ employ prompt or adapter tuning; all others are evaluated without domain-specific finetuning.}
    \label{tab:cuva_metrics}
\end{table}

\begin{table*}
    \centering
    \renewcommand{\arraystretch}{1.1}
    \resizebox{0.9\textwidth}{!}{\small
    \begin{tabular}{l ccc ccc ccc ccc}
        \toprule
        \textbf{Method} & \multicolumn{3}{c}{\textbf{BLEU}} & \multicolumn{3}{c}{\textbf{CIDEr}} & \multicolumn{3}{c}{\textbf{METEOR}} & \multicolumn{3}{c}{\textbf{ROUGE}} \\
        \cmidrule(lr){2-4} \cmidrule(lr){5-7} \cmidrule(lr){8-10} \cmidrule(lr){11-13}
        & C & E & V & C & E & V & C & E & V & C & E & V \\
        \midrule
        Video-ChatGPT \cite{Maaz2023VideoChatGPT} & 0.152 & 0.068 & 0.066 & 0.033 & 0.011 & 0.013 & 0.102 & 0.069 & 0.044 & 0.153 & 0.048 & 0.079 \\
        Video-LLAMA \cite{damonlpsg2023videollama} & 0.151 & 0.079 & 0.104 & 0.024 & 0.014 & 0.017 & 0.112 & 0.076 & 0.057 & 0.156 & 0.067 & 0.090 \\
        Video-LLAVA \cite{lin2023video} & 0.164 & 0.046 & 0.055 & 0.032 & 0.009 & 0.013 & 0.097 & 0.022 & 0.014 & 0.132 & 0.023 & 0.045 \\
        LLAVA-Next-Video \cite{zhang2024videoinstructiontuningsynthetic} & 0.435 & 0.091 & 0.120 & 0.102 & 0.015 & 0.031 & 0.117 & 0.085 & 0.096 & 0.198 & 0.080 & 0.106 \\
        QwenVL2 \cite{Qwen2-VL} & 0.312 & 0.082 & 0.155 & 0.044 & 0.020 & 0.044 & 0.133 & 0.092 & 0.112 & 0.163 & 0.081 & 0.137 \\
        InternVL2 \cite{chen2024far} & 0.331 & 0.101 & 0.145 & 0.052 & 0.022 & 0.035 & 0.141 & 0.095 & 0.101 & 0.182 & 0.102 & 0.122 \\
        NVILA \cite{li2025otter} & 0.610 & 0.340 & 0.283 & 0.261 & 0.154 & 0.098 & 0.157 & 0.096 & 0.072 & 0.273 & 0.218 & 0.198 \\
        Holmes-VAU$^{\ddag}$ \cite{zhang2025holmes} & {0.913} & {0.804} & {0.566} & {0.467} & {1.519} & \underline{1.437} & {0.190} & {0.165} & {0.121} & {0.329} & {0.370} & {0.355} \\
        \rowcolor{SoftBlue}{VADER$^{\ddag}$ (Ours)} & \underline{1.035} & \underline{1.163} & \underline{1.068} & \underline{0.612} & \underline{1.650} & {1.403} & \underline{0.215} & \underline{0.183} & \underline{0.137} & \underline{0.376} & \underline{0.444} & \underline{0.408} \\
        \rowcolor{SoftBlue}\textbf{VADER$^{\dag}$ (Ours)} & \textbf{1.266} & \textbf{1.246} & \textbf{1.268} & \textbf{1.040} & \textbf{1.763} & \textbf{1.812} & \textbf{0.247} & \textbf{0.216} & \textbf{0.164} & \textbf{0.429} & \textbf{0.463} & \textbf{0.446} \\
        \bottomrule
    \end{tabular}}
    \caption{\textbf{Evaluation on the HIVAU-70k benchmark.} The dataset supports hierarchical evaluation at the clip (C), event (E), and video (V) levels. VADER achieves consistently strong results across metrics and levels. Models marked with $^{\dag}$ are finetuned on HIVAU-70k; those marked with $^{\ddag}$ are finetuned and use the same backbone size as baseline~\cite{zhang2025holmes}; all others are evaluated without domain-specific finetuning.}
    \label{tab:hivau}
\end{table*}

\renewcommand{\arraystretch}{1.2}
\begin{table*}[t]
    \centering
    \resizebox{\textwidth}{!}{
    \begin{tabular}{l cccc ccc | cccc ccc}
        \toprule
        \multicolumn{8}{c|}{\textbf{Anomaly Video Description Generation}} & \multicolumn{7}{c}{\textbf{Anomaly Video Question-Answering}} \\
        \midrule
        & \multicolumn{4}{c}{\textbf{Text-Level}} & \multicolumn{3}{c|}{\textbf{GPT-Guided}} & \multicolumn{4}{c}{\textbf{Text-Level}} & \multicolumn{3}{c}{\textbf{GPT-Guided}} \\
        \cmidrule(lr){2-5} \cmidrule(lr){6-8} \cmidrule(lr){9-12} \cmidrule(lr){13-15}
        \textbf{Method} & BLEU-1 & BLEU-2 & BLEU-3 & BLEU-4 & Reason. & Detail & Consist. & BLEU-1 & BLEU-2 & BLEU-3 & BLEU-4 & Reason. & Detail & Consist. \\
        \midrule
        Video-ChatGPT \cite{Maaz2023VideoChatGPT} & 0.107 & 0.046 & 0.017 & 0.008 & 0.084 & 0.108 & 0.055 & 0.177 & 0.096 & 0.058 & 0.038 & 0.508 & 0.430 & 0.421 \\
        VideoChat \cite{li2023videochat} & 0.053 & 0.023 & 0.008 & 0.003 & 0.107 & 0.205 & 0.054 & 0.261 & 0.133 & 0.074 & 0.043 & 0.699 & 0.631 & 0.598 \\
        Video-LLAMA \cite{damonlpsg2023videollama} & 0.062 & 0.025 & 0.009 & 0.004 & 0.120 & 0.217 & 0.066 & 0.156 & 0.081 & 0.045 & 0.027 & 0.586 & 0.485 & 0.497 \\
        LLAMA-Adapter \cite{zhang2023llamaadapter} & 0.132 & 0.052 & 0.018 & 0.008 & 0.060 & 0.091 & 0.038 & 0.199 & 0.109 & 0.067 & 0.043 & 0.646 & 0.559 & 0.549 \\
        Video-LLAVA \cite{lin2023video} & 0.071 & 0.030 & 0.012 & 0.005 & 0.077 & 0.115 & 0.038 & 0.094 & 0.054 & 0.034 & 0.023 & 0.393 & 0.274 & 0.316 \\
        NVILA \cite{li2025otter} & 0.132 & 0.058 & 0.016 & 0.003 & 0.355 & \textbf{0.527} & \underline{0.299} & 0.003 & 0.001 & 0.001 & 0.001 & 0.428 & 0.304 & 0.362 \\
        HAWK$^{\dag}$ \cite{atang2024hawk} & \underline{0.270} & \underline{0.139} & \underline{0.074} & \underline{0.043} & \underline{0.283} & 0.320 & 0.218 & \underline{0.319} & \underline{0.179} & \underline{0.112} & \underline{0.073} & \textbf{0.840} & \underline{0.794} & \underline{0.753} \\
        \rowcolor{SoftBlue}\textbf{VADER$^{\dag}$ (Ours)} & \textbf{0.324} & \textbf{0.196} & \textbf{0.127} & \textbf{0.071} & \textbf{0.428} & \underline{0.442} & \textbf{0.357} & \textbf{0.484} & \textbf{0.311} & \textbf{0.210} & \textbf{0.150} & \underline{0.828} & \textbf{0.825} & \textbf{0.794} \\
        \bottomrule
    \end{tabular}
    }
    \caption{\textbf{Evaluation on the HAWK benchmark.} VADER achieves leading results in both anomaly description generation and question-answering tasks across text-level and GPT-guided metrics. \textbf{Bold} indicate the best performance, and \textit{underlined} indicate the second best. Models marked with $^{\dag}$ are finetuned on HAWK; all others are evaluated without domain-specific finetuning.}
    \label{tab:hawk_results}
\end{table*}

\subsection{Benchmark Datasets and Evaluation Metrics}

To evaluate the validity of VADER, we use three latest comprehensive VAU benchmarks: HIVAU-70k \cite{zhang2025holmes}, HAWK \cite{atang2024hawk}, and CUVA\cite{CUVA}.

\textbf{HIVAU-70k}~\cite{zhang2025holmes} is a large-scale dataset comprising videos from UCF-Crime~\cite{wu2020not} and XD-Violence~\cite{sultani2018real}, providing hierarchical annotations at clip, event, and video levels for both perception and reasoning tasks. 

\textbf{HAWK}~\cite{atang2024hawk} focuses on open-world video anomaly understanding, collecting videos from seven diverse datasets~\cite{acsintoae2022ubnormal, chan2008modeling, liu2018future, lu2013abnormal, sultani2018real, wang2010anomaly, yao2022dota}, and supports both event description and question answering tasks. 

\textbf{CUVA}~\cite{CUVA} targets causation understanding in real-world anomaly videos, offering detailed annotations that explain what happened, why, and how for each event.

We adopt the evaluation metrics following the official protocols of each benchmark. These can be categorized into three groups: \textbf{(1) Lexical-level metrics}, including BLEU~\cite{papineni2002bleu}, ROUGE~\cite{lin2004rouge}, METEOR~\cite{banerjee2005meteor}, and CIDEr~\cite{vedantam2015cider}, measure n-gram overlap, precision, recall, and synonym matching with reference texts. \textbf{(2) Semantic-level metrics}, such as BLEURT~\cite{sellam2020bleurt} and MoverScore~\cite{zhao2019moverscore}, use pretrained language models to capture meaning and contextual similarity beyond exact word matches. \textbf{(3) Judge-based evaluative metrics}, including GPT score~\cite{achiam2023gpt} and MMEval~\cite{CUVA}, directly use LLMs or VLMs as evaluators to judge plausibility, informativeness, and consistency of the generated explanations with video content, simulating human judgment in open-ended settings. 

All metrics are reported such that higher scores indicate better performance. Table~\ref{tab:metrics_summary} summarizes the metrics used for each benchmark.

\subsection{Experimental Results}
We present quantitative evaluations of VADER across three challenging benchmarks, highlighting its robust and superior performance.

\textbf{Performance on CUVA.}
Table~\ref{tab:cuva_metrics} demonstrates VADER's competitive performance across anomaly description and causation tasks.
VADER achieves the highest MMEval score for "Causes" task with an improvement of 7.38 points and maintains a comparable score on the other tasks. While VADER’s BLEURT scores are slightly lower than some baselines, this is due to BLEURT’s focus on surface-level wording similarity. In contrast, VADER excels on human-aligned metrics such as UniEval and MMEval, indicating its strength in generating coherent and accurate causal explanations beyond lexical overlap.

\textbf{Performance on HIVAU-70k.}
As shown in Table~\ref{tab:hivau}, VADER achieves state-of-the-art results across all metrics and evaluation levels (clip, event, and video). The performance gains are particularly significant at the event and video levels, highlighting VADER’s ability to reason over complex, temporally extended anomalous events and generate coherent, causally-aware narratives.

\textbf{Performance on HAWK.}
Results presented in Table~\ref{tab:hawk_results} illustrate VADER’s strong performance in anomaly description and question-answering tasks. For description generation, VADER surpasses previous best models by 0.045 in BLEU-1 and 0.026 in BLEU-4. For the question-answering task, VADER demonstrates even greater improvements, outperforming prior methods by 0.164 in BLEU-1 and 0.077 in BLEU-4. Additionally, it achieves leading GPT-guided scores, notably in Detail (description task) and Consistency (QA task), underscoring its ability to generate detailed, plausible, and contextually consistent narratives.

These results underscore VADER's versatility and high performance across diverse evaluation scenarios.

\subsection{Ablation Study}
We conduct ablation studies on the HAWK~\cite{atang2024hawk} description task to systematically assess the contribution of each component in VADER.

\textbf{Impact of Key Components.}
As shown in Table~\ref{tab:ablation_components}, removing the relation reasoning branch, CORE, leads to a consistent drop in both text-level and GPT-guided metrics, confirming the importance of modeling object interactions. Disabling the proposed context-aware sampling strategy, CAES, also degrades performance across all metrics, further validating its effectiveness. Without fine-tuning, performance drops significantly, though the GPT-guided detail score is anomalously high, likely because the model generates lengthy but less focused responses due to the lack of domain adaptation.

\begin{table*}[ht!] 
    \centering
    \renewcommand{\arraystretch}{1.2}

    \begin{minipage}[t]{0.48\textwidth}
        \centering
        \resizebox{\linewidth}{!}{
        \begin{tabular}{l cc ccc}
        \toprule
        & \multicolumn{2}{c}{\textbf{Text-Level}} & \multicolumn{3}{c}{\textbf{GPT-Guided}} \\
        \cmidrule(lr){2-3} \cmidrule(lr){4-6}
        \textbf{Method} & BLEU & ROUGE & Reasonability & Detail & Consistency \\
        \midrule
        \rowcolor{SoftBlue}\textbf{VADER} & \textbf{0.718} & \textbf{0.283} & \textbf{0.428} & 0.442 & \textbf{0.357} \\
        w/o CORE & 0.668 & 0.274 & 0.419 & 0.477 & 0.343 \\
        w/o CAES & 0.594 & 0.244 & 0.387 & 0.435 & 0.331 \\
        w/o Fine-Tuning & 0.209 & 0.143 & 0.355 & \textbf{0.527} & 0.299 \\
        \bottomrule
    \end{tabular}}
    \caption{\textbf{Ablation study on key components of VADER.} Removing either CORE, CAES, or fine-tuning leads to performance drops in both text-level and GPT-guided metrics, underscoring the importance of each module to the overall performance.}
    \label{tab:ablation_components}
    \end{minipage}%
    \hfill 
    \begin{minipage}[t]{0.48\textwidth}
        \centering
        \resizebox{\linewidth}{!}{
        \begin{tabular}{l cc ccc}
        \toprule
        & \multicolumn{2}{c}{\textbf{Text-Level}} & \multicolumn{3}{c}{\textbf{GPT-Guided}} \\
        \cmidrule(lr){2-3} \cmidrule(lr){4-6}
        \textbf{Method} & BLEU & ROUGE & Reasonability & Detail & Consistency \\
        \midrule
        Uniform & 0.594 & 0.244 & 0.387 & 0.435 & 0.331 \\
        Top-K & 0.663 & 0.273 & 0.381 & 0.411 & 0.327 \\
        ATS~\cite{zhang2025holmes} & 0.641 & 0.273 & 0.383 & 0.423 & 0.335 \\
        \rowcolor{SoftBlue}\textbf{CAES (Ours)} & \textbf{0.668} & \textbf{0.274} & \textbf{0.419} & \textbf{0.477} & \textbf{0.343} \\
        \bottomrule
    \end{tabular}}
    \caption{\textbf{Comparison of keyframe selection strategies.} Our Context-AwarE Sampling (CAES) outperforms uniform, Top-K, and ATS approaches across text-level and GPT-guided metrics. }
    \label{tab:ablation_kfselect}
    \end{minipage}

    \vspace{2em} 

    \begin{minipage}[t]{0.48\textwidth}
        \centering
        \resizebox{\linewidth}{!}{
        \begin{tabular}{l cc ccc}
        \toprule
        & \multicolumn{2}{c}{\textbf{Text-Level}} & \multicolumn{3}{c}{\textbf{GPT-Guided}} \\
        \cmidrule(lr){2-3} \cmidrule(lr){4-6}
        \textbf{Context Sampling Method} & BLEU & ROUGE & Reasonability & Detail & Consistency \\
        \midrule
        Fixed window & 0.647 & 0.273 & 0.419 & 0.477 & 0.343 \\
        Exponential interval & 0.665 & \textbf{0.278} & 0.392 & 0.421 & 0.342 \\
        \rowcolor{SoftBlue}\textbf{Dynamic window} & \textbf{0.668} & 0.274 & \textbf{0.419} & \textbf{0.477} & \textbf{0.343} \\
        \bottomrule
    \end{tabular}}
    \caption{\textbf{Ablation study on context sampling strategies in CAES.} Dynamic window sampling achieves the best overall performance across both text-level and GPT-guided metrics. }
    \label{tab:ablation_context}
    \end{minipage}%
    \hfill 
    \begin{minipage}[t]{0.48\textwidth}
        \centering
        \resizebox{\linewidth}{!}{
        \begin{tabular}{l cc ccc}
        \toprule
        & \multicolumn{2}{c}{\textbf{Text-Level}} & \multicolumn{3}{c}{\textbf{GPT-Guided}} \\
        \cmidrule(lr){2-3} \cmidrule(lr){4-6}
        \textbf{Method} & BLEU & ROUGE & Reasonability & Detail & Consistency \\
        \midrule
        Relational visual cue & 0.670 & 0.276 & 0.385 & 0.430 & 0.326 \\
        Scene graph text & 0.686 & 0.277 & 0.391 & 0.434 & 0.335 \\
        \rowcolor{SoftBlue}\textbf{Relation token} & \textbf{0.718} & \textbf{0.283} & \textbf{0.428} & \textbf{0.442} & \textbf{0.357} \\
        \bottomrule
    \end{tabular}}
    \caption{\textbf{Comparison of relation representations.} Using relation tokens as relation representations outperforms both relational visual cue and scene graph text approaches across all text-level and GPT-guided metrics.}
    \label{tab:ablation_relation_repr}
    \end{minipage}
    
\end{table*}

\begin{figure}[h!]
    \centering
    \includegraphics[width=0.48\textwidth]{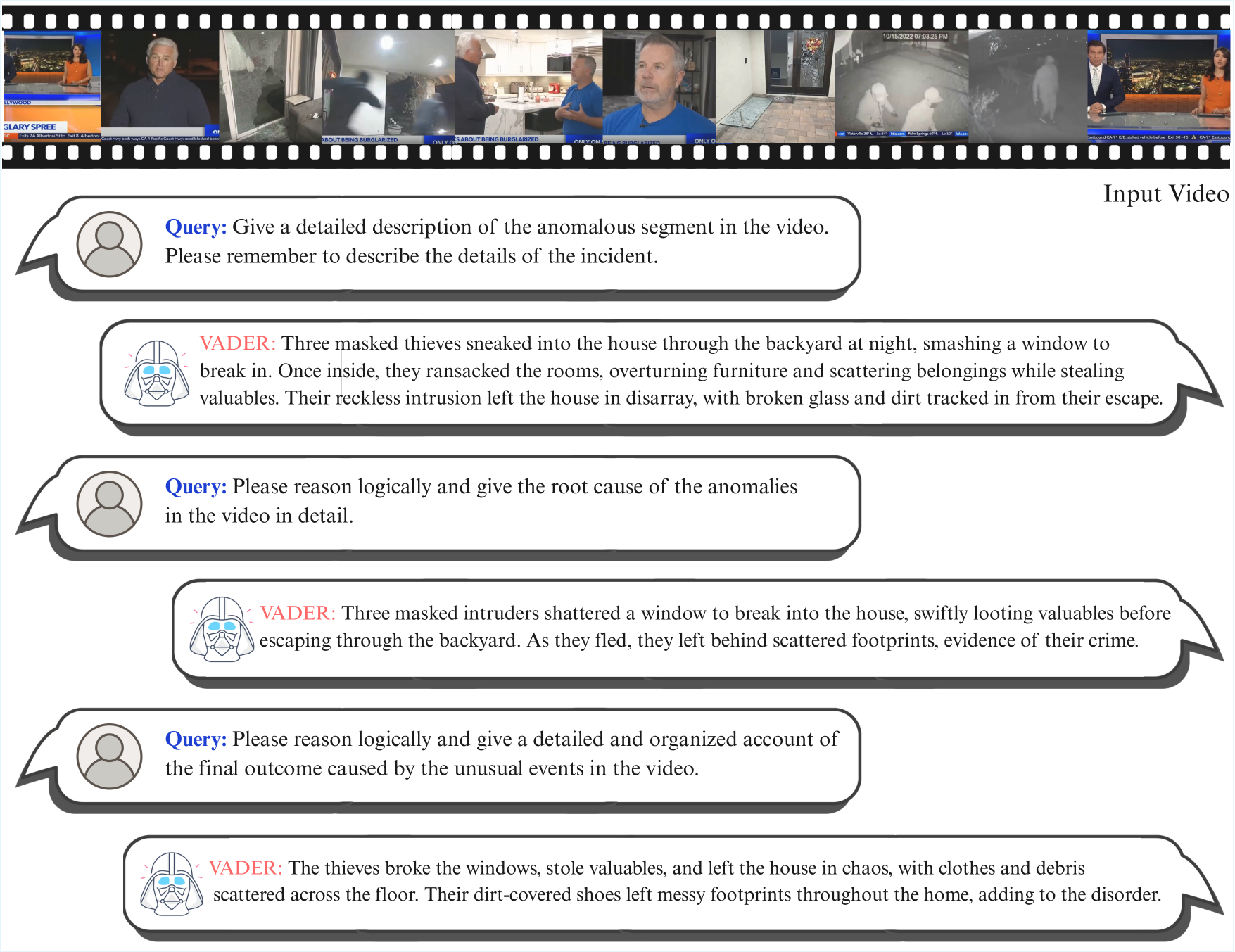}
    \caption{\textbf{Qualitative results of VADER's causal reasoning capabilities.} Given a video depicting a nighttime break-in, VADER generates detailed, context-aware answers for a sequence of causal reasoning queries, capturing both the key actions and underlying cause-effect chains within the event.}
    \label{fig:causal}
\end{figure}

\textbf{Effect of Sampling Strategy.}
Table~\ref{tab:ablation_kfselect} compares our CAES with alternative keyframe selection methods. Uniform sampling results in lower BLEU and ROUGE scores, as it may overlook key anomaly frames. Top-K focuses on frames with the highest anomaly scores, often over-concentrating on anomalous segments and neglecting essential context. The ATS method~\cite{zhang2025holmes} adaptively samples frames based on anomaly scores using a density-aware approach; however, in our setting, it shows only marginal gains over Top-K on text-level metrics, likely due to limited context diversity in complex scenarios. Our CAES achieves the best trade-off across both text-level and GPT-guided metrics, effectively balancing anomaly saliency and context diversity to produce coherent event narratives.

\textbf{Effect of Context Sampling Strategies.} 
Table~\ref{tab:ablation_context} compares different approaches for selecting pre/post-event context frames in CAES. Fixed Window samples consecutive frames immediately before the anomaly, which may lead to high redundancy and limited temporal scope. Exponential Interval samples at increasing intervals to provide multi-scale context, but may still miss critical transitions. Our Dynamic Window adaptively determines context boundaries based on anomaly score gradients, enabling event-specific, informative context coverage. The results show that Dynamic Window achieves strong performance across metrics, highlighting the value of adaptive, narrative-aware context selection.

\textbf{Relation Representation.}
Table~\ref{tab:ablation_relation_repr} compares different strategies for encoding relational information. Using only relational visual cues encodes the detected objects’ static visual features and locations, but fails to capture interactions or evolving relationships between objects. Scene graph text descriptions leverage the relationship labels predicted by the pretrained EGTR~\cite{im2024egtr} on Visual Genome~\cite{krishna2017visual}, but these generic labels are not tailored for the nuanced, context-specific interactions common in anomaly scenarios. In contrast, our relation token, generated by CORE, provides continuous and fine-grained temporal cues, yielding consistently better results across all metrics.

\begin{figure*}[ht!]
    \centering
    \includegraphics[width=\textwidth]{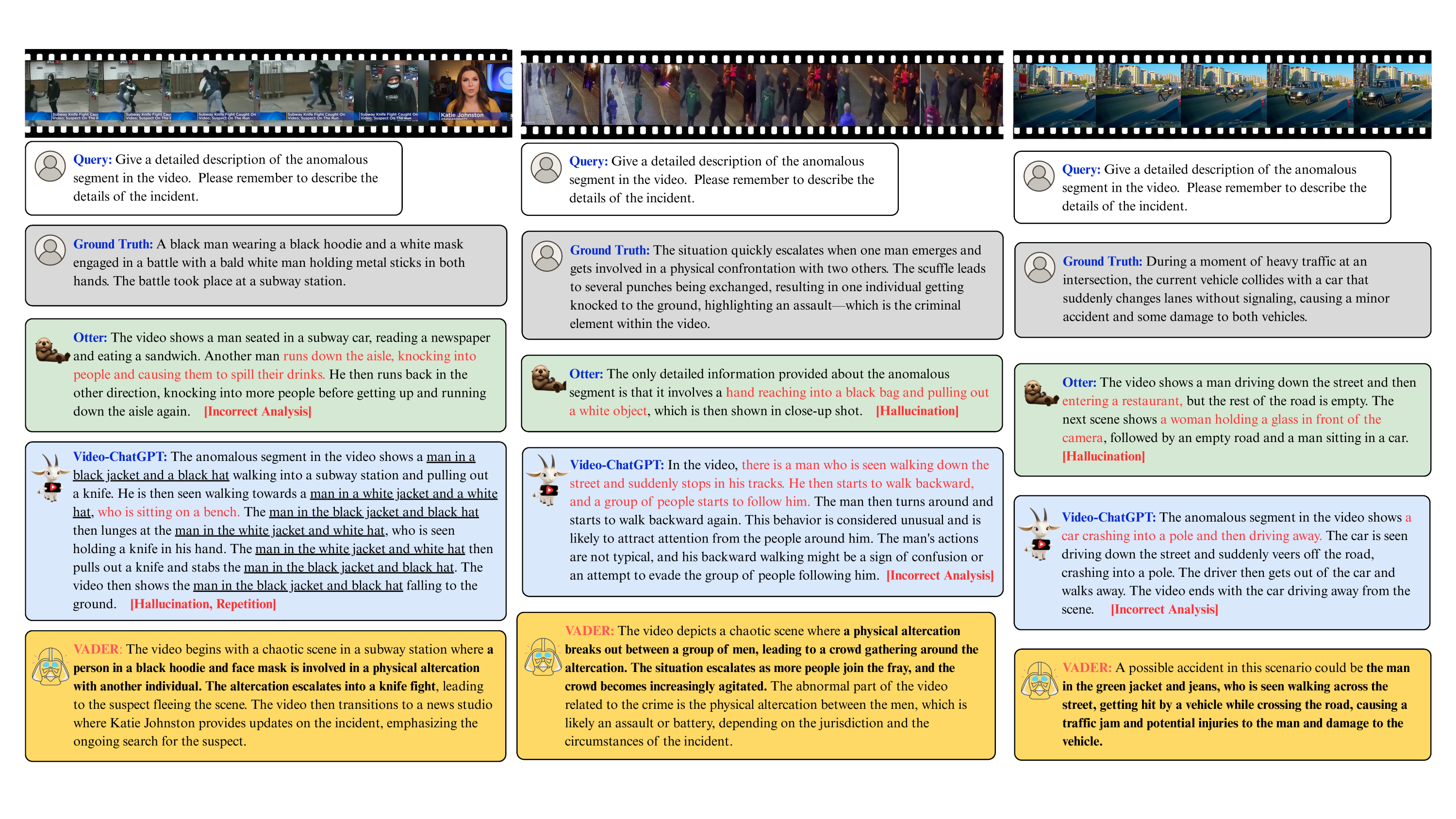}
    \caption{
    Three examples of the task of describing anonymous videos are depicted here. The descriptions generated by Otter ~\cite{li2025otter} and Video-ChatGPT~\cite{Maaz2023VideoChatGPT} contain hallucination or incorrect analysis. In contrast, VADER produces concise, contextually grounded, and causally coherent descriptions that accurately reflect the events and their underlying dynamics across various challenging cases.}
    \label{fig:visualization}
\end{figure*}

\subsection{Qualitative Results}
Figure~\ref{fig:causal} illustrates a qualitative example that highlights VADER's strengths in causal reasoning. Given a video of a nighttime break-in, VADER offers a detailed description, root cause, and outcome summary of the anomalous event. Its responses capture not only key visual cues such as the broken window and scattered belongings, but also the underlying object-level interactions, such as the intruders' actions in the environment, demonstrating the benefit of our keyframe selection and relational modeling.

Figure~\ref{fig:visualization} further compares VADER with existing MLLMs~\cite{li2025otter, Maaz2023VideoChatGPT} on diverse real-world anomalous videos. Across various scenarios, physical altercations, road accidents, and complex crowd behaviors, VADER produces more accurate, context-aware, and semantically coherent descriptions than recent baselines. These results illustrate VADER’s ability to mitigate hallucinations, capture causal structure, and provide comprehensive explanations, thereby advancing explainable video anomaly understanding.

We provide additional qualitative results and failure cases in the Supp. Material Sec.~\ref{sec:supp_intermediate_vis} and~\ref{sec:supp_falure_case}.
\section{Limitation}
\label{sec:limitation}

While VADER advances video anomaly understanding, several limitations remain:

\textbf{Dependency on Upstream Models.} 
VADER relies on upstream modules such as the Anomaly Scorer and Relation Feature Extractor, where errors in detection or association can propagate and affect the final reasoning results.

\textbf{Inherent Bias towards High-Motion Events.}
The reliance on relational volatility as the core signal introduces a bias toward anomalies with strong motion, making the framework less sensitive to subtle or low-motion anomalies. 

\textbf{Limitation to Object-Centric View.}
VADER’s current design is object-centric, which limits its ability to capture scene-level or group-level anomalies such as environmental changes or collective crowd behaviors. 

Detailed potential mitigation strategies are discussed in Supp. Material Sec.~\ref{sec:supp_mitigation_for_limitations}.
\section{Conclusion}

In this work, we presented VADER, a novel framework for video anomaly understanding that leverages Context-AwarE Sampling (CAES) and dynamic relational modeling to deliver detailed, causally grounded interpretations of anomalous events. Extensive experiments on multiple challenging benchmarks demonstrate that VADER achieves state-of-the-art or highly competitive performance across anomaly description, causal explanation, and question answering tasks. Systematic ablation studies further highlight the critical roles of adaptive context sampling and relational dynamics encoding.

\newpage
\noindent\textbf{Acknowledgement} 
This work is supported by the NVIDIA Taiwan AI Research \& Development Center (TRDC).
{
    \small
    \bibliographystyle{ieeenat_fullname}
    \bibliography{main}

\begin{thebibliography}{74}
\providecommand{\natexlab}[1]{#1}
\providecommand{\url}[1]{\texttt{#1}}
\expandafter\ifx\csname urlstyle\endcsname\relax
  \providecommand{\doi}[1]{doi: #1}\else
  \providecommand{\doi}{doi: \begingroup \urlstyle{rm}\Url}\fi

\bibitem[Achiam et~al.(2023)Achiam, Adler, Agarwal, Ahmad, Akkaya, Aleman, Almeida, Altenschmidt, Altman, Anadkat, et~al.]{achiam2023gpt}
Josh Achiam, Steven Adler, Sandhini Agarwal, Lama Ahmad, Ilge Akkaya, Florencia~Leoni Aleman, Diogo Almeida, Janko Altenschmidt, Sam Altman, Shyamal Anadkat, et~al.
\newblock Gpt-4 technical report.
\newblock \emph{arXiv preprint arXiv:2303.08774}, 2023.

\bibitem[Acsintoae et~al.(2022)Acsintoae, Florescu, Georgescu, Mare, Sumedrea, Ionescu, Khan, and Shah]{acsintoae2022ubnormal}
Andra Acsintoae, Andrei Florescu, Mariana-Iuliana Georgescu, Tudor Mare, Paul Sumedrea, Radu~Tudor Ionescu, Fahad~Shahbaz Khan, and Mubarak Shah.
\newblock Ubnormal: New benchmark for supervised open-set video anomaly detection.
\newblock In \emph{Proceedings of the IEEE/CVF conference on computer vision and pattern recognition}, pages 20143--20153, 2022.

\bibitem[Alayrac et~al.(2022)Alayrac, Donahue, Luc, Miech, Barr, Hasson, Lenc, Mensch, Millican, Reynolds, et~al.]{alayrac2022flamingo}
Jean-Baptiste Alayrac, Jeff Donahue, Pauline Luc, Antoine Miech, Iain Barr, Yana Hasson, Karel Lenc, Arthur Mensch, Katherine Millican, Malcolm Reynolds, et~al.
\newblock Flamingo: a visual language model for few-shot learning.
\newblock \emph{Advances in neural information processing systems}, 35:\penalty0 23716--23736, 2022.

\bibitem[Banerjee and Lavie(2005)]{banerjee2005meteor}
Satanjeev Banerjee and Alon Lavie.
\newblock Meteor: An automatic metric for mt evaluation with improved correlation with human judgments.
\newblock In \emph{Proceedings of the acl workshop on intrinsic and extrinsic evaluation measures for machine translation and/or summarization}, pages 65--72, 2005.

\bibitem[Cao et~al.(2025)Cao, Xu, Cheng, Sun, Du, Gao, and Shen]{cao2025personalizing}
Yunkang Cao, Xiaohao Xu, Yuqi Cheng, Chen Sun, Zongwei Du, Liang Gao, and Weiming Shen.
\newblock Personalizing vision-language models with hybrid prompts for zero-shot anomaly detection.
\newblock \emph{IEEE Transactions on Cybernetics}, 2025.

\bibitem[Carion et~al.(2020)Carion, Massa, Synnaeve, Usunier, Kirillov, and Zagoruyko]{carion2020end}
Nicolas Carion, Francisco Massa, Gabriel Synnaeve, Nicolas Usunier, Alexander Kirillov, and Sergey Zagoruyko.
\newblock End-to-end object detection with transformers.
\newblock In \emph{European conference on computer vision}, pages 213--229. Springer, 2020.

\bibitem[Chan and Vasconcelos(2008)]{chan2008modeling}
Antoni~B Chan and Nuno Vasconcelos.
\newblock Modeling, clustering, and segmenting video with mixtures of dynamic textures.
\newblock \emph{IEEE transactions on pattern analysis and machine intelligence}, 30\penalty0 (5):\penalty0 909--926, 2008.

\bibitem[Chen et~al.(2025)Chen, Liu, Song, Li, Yuan, Yu, and Pang]{chen2025unveiling}
Chenglizhao Chen, Xinyu Liu, Mengke Song, Luming Li, Shaojiang Yuan, Xu Yu, and Shanchen Pang.
\newblock Unveiling context-related anomalies: Knowledge graph empowered decoupling of scene and action for human-related video anomaly detection.
\newblock \emph{IEEE Transactions on Circuits and Systems for Video Technology}, 2025.

\bibitem[Chen et~al.(2024)Chen, Wang, Tian, Ye, Gao, Cui, Tong, Hu, Luo, Ma, et~al.]{chen2024far}
Zhe Chen, Weiyun Wang, Hao Tian, Shenglong Ye, Zhangwei Gao, Erfei Cui, Wenwen Tong, Kongzhi Hu, Jiapeng Luo, Zheng Ma, et~al.
\newblock How far are we to gpt-4v? closing the gap to commercial multimodal models with open-source suites.
\newblock \emph{Science China Information Sciences}, 67\penalty0 (12):\penalty0 220101, 2024.

\bibitem[Chong and Tay(2017)]{chong2017abnormal}
Yong~Shean Chong and Yong~Haur Tay.
\newblock Abnormal event detection in videos using spatiotemporal autoencoder.
\newblock In \emph{International symposium on neural networks}, pages 189--196. Springer, 2017.

\bibitem[Du et~al.(2024)Du, Zhang, Xie, Nan, Zhang, Xu, Liu, Leng, Liu, Fan, Huang, Feng, Chen, Zhang, Li, Zhang, Chen, Cui, and Tao]{CUVA}
Hang Du, Sicheng Zhang, Binzhu Xie, Guoshun Nan, Jiayang Zhang, Junrui Xu, Hangyu Liu, Sicong Leng, Jiangming Liu, Hehe Fan, Dajiu Huang, Jing Feng, Linli Chen, Can Zhang, Xuhuan Li, Hao Zhang, Jianhang Chen, Qimei Cui, and Xiaofeng Tao.
\newblock Uncovering what, why and how: A comprehensive benchmark for causation understanding of video anomaly.
\newblock In \emph{2024 IEEE/CVF Conference on Computer Vision and Pattern Recognition (CVPR)}, pages 18793--18803, 2024.

\bibitem[Gao et~al.(2025)Gao, Yang, and Huang]{gao2025suvad}
Shibo Gao, Peipei Yang, and Linlin Huang.
\newblock Suvad: Semantic understanding based video anomaly detection using mllm.
\newblock In \emph{ICASSP 2025-2025 IEEE International Conference on Acoustics, Speech and Signal Processing (ICASSP)}, pages 1--5. IEEE, 2025.

\bibitem[Georgescu et~al.(2021)Georgescu, Barbalau, Ionescu, Khan, Popescu, and Shah]{georgescu2021anomaly}
Mariana-Iuliana Georgescu, Antonio Barbalau, Radu~Tudor Ionescu, Fahad~Shahbaz Khan, Marius Popescu, and Mubarak Shah.
\newblock Anomaly detection in video via self-supervised and multi-task learning.
\newblock In \emph{Proceedings of the IEEE/CVF conference on computer vision and pattern recognition}, pages 12742--12752, 2021.

\bibitem[Hasan et~al.(2016)Hasan, Choi, Neumann, Roy-Chowdhury, and Davis]{hasan2015context}
Mahmudul Hasan, Jonghyun Choi, jan Neumann, Amit~K Roy-Chowdhury, and Larry Davis.
\newblock Learning temporal regularity in video sequences.
\newblock In \emph{Proceedings of IEEE Computer Vision and Pattern Recognition}, 2016.

\bibitem[Hu et~al.(2022)Hu, Shen, Wallis, Allen-Zhu, Li, Wang, Wang, Chen, et~al.]{hu2022lora}
Edward~J Hu, Yelong Shen, Phillip Wallis, Zeyuan Allen-Zhu, Yuanzhi Li, Shean Wang, Lu Wang, Weizhu Chen, et~al.
\newblock Lora: Low-rank adaptation of large language models.
\newblock \emph{ICLR}, 1\penalty0 (2):\penalty0 3, 2022.

\bibitem[Hu et~al.(2024)Hu, Tu, Han, He, Cui, Long, Zheng, Fang, Huang, Zhao, et~al.]{hu2024minicpm}
Shengding Hu, Yuge Tu, Xu Han, Chaoqun He, Ganqu Cui, Xiang Long, Zhi Zheng, Yewei Fang, Yuxiang Huang, Weilin Zhao, et~al.
\newblock Minicpm: Unveiling the potential of small language models with scalable training strategies.
\newblock \emph{arXiv preprint arXiv:2404.06395}, 2024.

\bibitem[Im et~al.(2024)Im, Nam, Park, Lee, and Park]{im2024egtr}
Jinbae Im, JeongYeon Nam, Nokyung Park, Hyungmin Lee, and Seunghyun Park.
\newblock Egtr: Extracting graph from transformer for scene graph generation.
\newblock In \emph{Proceedings of the IEEE/CVF Conference on Computer Vision and Pattern Recognition}, pages 24229--24238, 2024.

\bibitem[Krishna et~al.(2017)Krishna, Zhu, Groth, Johnson, Hata, Kravitz, Chen, Kalantidis, Li, Shamma, et~al.]{krishna2017visual}
Ranjay Krishna, Yuke Zhu, Oliver Groth, Justin Johnson, Kenji Hata, Joshua Kravitz, Stephanie Chen, Yannis Kalantidis, Li-Jia Li, David~A Shamma, et~al.
\newblock Visual genome: Connecting language and vision using crowdsourced dense image annotations.
\newblock \emph{International journal of computer vision}, 123\penalty0 (1):\penalty0 32--73, 2017.

\bibitem[Kuhn(1955)]{kuhn1955hungarian}
Harold~W Kuhn.
\newblock The hungarian method for the assignment problem.
\newblock \emph{Naval research logistics quarterly}, 2\penalty0 (1-2):\penalty0 83--97, 1955.

\bibitem[Li et~al.(2025)Li, Zhang, Chen, Wang, Pu, Cahyono, Yang, Li, and Liu]{li2025otter}
Bo Li, Yuanhan Zhang, Liangyu Chen, Jinghao Wang, Fanyi Pu, Joshua~Adrian Cahyono, Jingkang Yang, Chunyuan Li, and Ziwei Liu.
\newblock Otter: A multi-modal model with in-context instruction tuning.
\newblock \emph{IEEE Transactions on Pattern Analysis and Machine Intelligence}, 2025.

\bibitem[Li et~al.(2023{\natexlab{a}})Li, Li, Savarese, and Hoi]{li2023blip}
Junnan Li, Dongxu Li, Silvio Savarese, and Steven Hoi.
\newblock Blip-2: Bootstrapping language-image pre-training with frozen image encoders and large language models.
\newblock In \emph{International conference on machine learning}, pages 19730--19742. PMLR, 2023{\natexlab{a}}.

\bibitem[Li et~al.(2023{\natexlab{b}})Li, He, Wang, Li, Wang, Luo, Wang, Wang, and Qiao]{li2023videochat}
KunChang Li, Yinan He, Yi Wang, Yizhuo Li, Wenhai Wang, Ping Luo, Yali Wang, Limin Wang, and Yu Qiao.
\newblock Videochat: Chat-centric video understanding.
\newblock \emph{arXiv preprint arXiv:2305.06355}, 2023{\natexlab{b}}.

\bibitem[Li et~al.(2013)Li, Mahadevan, and Vasconcelos]{li2013anomaly}
Weixin Li, Vijay Mahadevan, and Nuno Vasconcelos.
\newblock Anomaly detection and localization in crowded scenes.
\newblock \emph{IEEE transactions on pattern analysis and machine intelligence}, 36\penalty0 (1):\penalty0 18--32, 2013.

\bibitem[Lin et~al.(2023)Lin, Zhu, Ye, Ning, Jin, and Yuan]{lin2023video}
Bin Lin, Bin Zhu, Yang Ye, Munan Ning, Peng Jin, and Li Yuan.
\newblock Video-llava: Learning united visual representation by alignment before projection.
\newblock \emph{arXiv preprint arXiv:2311.10122}, 2023.

\bibitem[Lin(2004)]{lin2004rouge}
Chin-Yew Lin.
\newblock Rouge: A package for automatic evaluation of summaries.
\newblock In \emph{Text summarization branches out}, pages 74--81, 2004.

\bibitem[Liu et~al.(2023{\natexlab{a}})Liu, Li, Wu, and Lee]{liu2023visual}
Haotian Liu, Chunyuan Li, Qingyang Wu, and Yong~Jae Lee.
\newblock Visual instruction tuning.
\newblock \emph{Advances in neural information processing systems}, 36:\penalty0 34892--34916, 2023{\natexlab{a}}.

\bibitem[Liu et~al.(2018)Liu, Luo, Lian, and Gao]{liu2018future}
Wen Liu, Weixin Luo, Dongze Lian, and Shenghua Gao.
\newblock Future frame prediction for anomaly detection--a new baseline.
\newblock In \emph{Proceedings of the IEEE conference on computer vision and pattern recognition}, pages 6536--6545, 2018.

\bibitem[Liu et~al.(2023{\natexlab{b}})Liu, Chang, Ma, Shan, and Chen]{liu2023diversity}
Wenrui Liu, Hong Chang, Bingpeng Ma, Shiguang Shan, and Xilin Chen.
\newblock Diversity-measurable anomaly detection.
\newblock In \emph{Proceedings of the IEEE/CVF conference on computer vision and pattern recognition}, pages 12147--12156, 2023{\natexlab{b}}.

\bibitem[Liu et~al.(2021)Liu, Nie, Long, Zhang, and Li]{liu2021hybrid}
Zhian Liu, Yongwei Nie, Chengjiang Long, Qing Zhang, and Guiqing Li.
\newblock A hybrid video anomaly detection framework via memory-augmented flow reconstruction and flow-guided frame prediction.
\newblock In \emph{Proceedings of the IEEE/CVF international conference on computer vision}, pages 13588--13597, 2021.

\bibitem[Liu et~al.(2024)Liu, Dong, Liu, Hu, Lu, and Rao]{liu2024oryx}
Zuyan Liu, Yuhao Dong, Ziwei Liu, Winston Hu, Jiwen Lu, and Yongming Rao.
\newblock Oryx mllm: On-demand spatial-temporal understanding at arbitrary resolution.
\newblock \emph{arXiv preprint arXiv:2409.12961}, 2024.

\bibitem[Liu et~al.(2025)Liu, Zhu, Shi, Zhang, Lou, Yang, Xi, Cao, Gu, Li, et~al.]{liu2025nvila}
Zhijian Liu, Ligeng Zhu, Baifeng Shi, Zhuoyang Zhang, Yuming Lou, Shang Yang, Haocheng Xi, Shiyi Cao, Yuxian Gu, Dacheng Li, et~al.
\newblock Nvila: Efficient frontier visual language models.
\newblock In \emph{Proceedings of the Computer Vision and Pattern Recognition Conference}, pages 4122--4134, 2025.

\bibitem[Lohner et~al.(2024)Lohner, Compagno, Francis, and Oltramari]{lohner2024enhancing}
Aaron Lohner, Francesco Compagno, Jonathan Francis, and Alessandro Oltramari.
\newblock Enhancing vision-language models with scene graphs for traffic accident understanding.
\newblock In \emph{2024 IEEE International Automated Vehicle Validation Conference (IAVVC)}, pages 1--7. IEEE, 2024.

\bibitem[Lu et~al.(2013)Lu, Shi, and Jia]{lu2013abnormal}
Cewu Lu, Jianping Shi, and Jiaya Jia.
\newblock Abnormal event detection at 150 fps in matlab.
\newblock In \emph{Proceedings of the IEEE international conference on computer vision}, pages 2720--2727, 2013.

\bibitem[Lv and Sun(2024)]{lv2024video}
Hui Lv and Qianru Sun.
\newblock Video anomaly detection and explanation via large language models.
\newblock \emph{arXiv preprint arXiv:2401.05702}, 2024.

\bibitem[Lv et~al.(2023)Lv, Yue, Sun, Luo, Cui, and Zhang]{lv2023unbiased}
Hui Lv, Zhongqi Yue, Qianru Sun, Bin Luo, Zhen Cui, and Hanwang Zhang.
\newblock Unbiased multiple instance learning for weakly supervised video anomaly detection.
\newblock In \emph{Proceedings of the IEEE/CVF conference on computer vision and pattern recognition}, pages 8022--8031, 2023.

\bibitem[Maaz et~al.(2024)Maaz, Rasheed, Khan, and Khan]{Maaz2023VideoChatGPT}
Muhammad Maaz, Hanoona Rasheed, Salman Khan, and Fahad~Shahbaz Khan.
\newblock Video-chatgpt: Towards detailed video understanding via large vision and language models.
\newblock In \emph{Proceedings of the 62nd Annual Meeting of the Association for Computational Linguistics (ACL 2024)}, 2024.

\bibitem[Mehran et~al.(2009)Mehran, Oyama, and Shah]{mehran2009abnormal}
Ramin Mehran, Alexis Oyama, and Mubarak Shah.
\newblock Abnormal crowd behavior detection using social force model.
\newblock In \emph{2009 IEEE conference on computer vision and pattern recognition}, pages 935--942. IEEE, 2009.

\bibitem[Nayak et~al.(2021)Nayak, Pati, and Das]{nayak2021comprehensive}
Rashmiranjan Nayak, Umesh~Chandra Pati, and Santos~Kumar Das.
\newblock A comprehensive review on deep learning-based methods for video anomaly detection.
\newblock \emph{Image and Vision Computing}, 106:\penalty0 104078, 2021.

\bibitem[Papineni et~al.(2002)Papineni, Roukos, Ward, and Zhu]{papineni2002bleu}
Kishore Papineni, Salim Roukos, Todd Ward, and Wei-Jing Zhu.
\newblock Bleu: a method for automatic evaluation of machine translation.
\newblock In \emph{Proceedings of the 40th annual meeting of the Association for Computational Linguistics}, pages 311--318, 2002.

\bibitem[Park et~al.(2020)Park, Noh, and Ham]{park2020learning}
Hyunjong Park, Jongyoun Noh, and Bumsub Ham.
\newblock Learning memory-guided normality for anomaly detection.
\newblock In \emph{Proceedings of the IEEE/CVF conference on computer vision and pattern recognition}, pages 14372--14381, 2020.

\bibitem[Radford et~al.(2021)Radford, Kim, Hallacy, Ramesh, Goh, Agarwal, Sastry, Askell, Mishkin, Clark, et~al.]{radford2021learning}
Alec Radford, Jong~Wook Kim, Chris Hallacy, Aditya Ramesh, Gabriel Goh, Sandhini Agarwal, Girish Sastry, Amanda Askell, Pamela Mishkin, Jack Clark, et~al.
\newblock Learning transferable visual models from natural language supervision.
\newblock In \emph{International conference on machine learning}, pages 8748--8763. PmLR, 2021.

\bibitem[Saligrama and Chen(2012)]{saligrama2012video}
Venkatesh Saligrama and Zhu Chen.
\newblock Video anomaly detection based on local statistical aggregates.
\newblock In \emph{2012 IEEE Conference on computer vision and pattern recognition}, pages 2112--2119. IEEE, 2012.

\bibitem[Schroff et~al.(2015)Schroff, Kalenichenko, and Philbin]{schroff2015facenet}
Florian Schroff, Dmitry Kalenichenko, and James Philbin.
\newblock Facenet: A unified embedding for face recognition and clustering.
\newblock In \emph{Proceedings of the IEEE conference on computer vision and pattern recognition}, pages 815--823, 2015.

\bibitem[Sellam et~al.(2020)Sellam, Das, and Parikh]{sellam2020bleurt}
Thibault Sellam, Dipanjan Das, and Ankur~P Parikh.
\newblock Bleurt: Learning robust metrics for text generation.
\newblock \emph{arXiv preprint arXiv:2004.04696}, 2020.

\bibitem[Su et~al.(2023)Su, Lan, Li, Xu, Wang, and Cai]{su2023pandagpt}
Yixuan Su, Tian Lan, Huayang Li, Jialu Xu, Yan Wang, and Deng Cai.
\newblock Pandagpt: One model to instruction-follow them all.
\newblock \emph{arXiv preprint arXiv:2305.16355}, 2023.

\bibitem[Sultani et~al.(2018)Sultani, Chen, and Shah]{sultani2018real}
Waqas Sultani, Chen Chen, and Mubarak Shah.
\newblock Real-world anomaly detection in surveillance videos.
\newblock In \emph{Proceedings of the IEEE conference on computer vision and pattern recognition}, pages 6479--6488, 2018.

\bibitem[Tang et~al.(2024)Tang, Lu, Wu, Xu, Ma, Fang, Guo, Lu, Chen, and Chen]{atang2024hawk}
Jiaqi Tang, Hao Lu, Ruizheng Wu, Xiaogang Xu, Ke Ma, Cheng Fang, Bin Guo, Jiangbo Lu, Qifeng Chen, and Ying-Cong Chen.
\newblock Hawk: Learning to understand open-world video anomalies.
\newblock In \emph{Neural Information Processing Systems (NeurIPS)}, 2024.

\bibitem[Tran et~al.(2015)Tran, Bourdev, Fergus, Torresani, and Paluri]{tran2015learning}
Du Tran, Lubomir Bourdev, Rob Fergus, Lorenzo Torresani, and Manohar Paluri.
\newblock Learning spatiotemporal features with 3d convolutional networks.
\newblock In \emph{Proceedings of the IEEE international conference on computer vision}, pages 4489--4497, 2015.

\bibitem[Vedantam et~al.(2015)Vedantam, Lawrence~Zitnick, and Parikh]{vedantam2015cider}
Ramakrishna Vedantam, C Lawrence~Zitnick, and Devi Parikh.
\newblock Cider: Consensus-based image description evaluation.
\newblock In \emph{Proceedings of the IEEE conference on computer vision and pattern recognition}, pages 4566--4575, 2015.

\bibitem[Wang et~al.(2024{\natexlab{a}})Wang, Qin, Bastola, Chen, Suchanek, Gong, and Razi]{wang2024visiongpt}
Hao Wang, Jiayou Qin, Ashish Bastola, Xiwen Chen, John Suchanek, Zihao Gong, and Abolfazl Razi.
\newblock Visiongpt: Llm-assisted real-time anomaly detection for safe visual navigation.
\newblock \emph{arXiv preprint arXiv:2403.12415}, 2024{\natexlab{a}}.

\bibitem[Wang et~al.(2024{\natexlab{b}})Wang, Bai, Tan, Wang, Fan, Bai, Chen, Liu, Wang, Ge, Fan, Dang, Du, Ren, Men, Liu, Zhou, Zhou, and Lin]{Qwen2-VL}
Peng Wang, Shuai Bai, Sinan Tan, Shijie Wang, Zhihao Fan, Jinze Bai, Keqin Chen, Xuejing Liu, Jialin Wang, Wenbin Ge, Yang Fan, Kai Dang, Mengfei Du, Xuancheng Ren, Rui Men, Dayiheng Liu, Chang Zhou, Jingren Zhou, and Junyang Lin.
\newblock Qwen2-vl: Enhancing vision-language model's perception of the world at any resolution.
\newblock \emph{arXiv preprint arXiv:2409.12191}, 2024{\natexlab{b}}.

\bibitem[Wang and Miao(2010)]{wang2010anomaly}
Shu Wang and Zhenjiang Miao.
\newblock Anomaly detection in crowd scene.
\newblock In \emph{IEEE 10th International Conference on Signal Processing Proceedings}, pages 1220--1223. IEEE, 2010.

\bibitem[Wang et~al.(2021)Wang, Che, Jiang, Xiao, Yang, Tang, Ye, Wang, and Qi]{wang2021robust}
Xuanzhao Wang, Zhengping Che, Bo Jiang, Ning Xiao, Ke Yang, Jian Tang, Jieping Ye, Jingyu Wang, and Qi Qi.
\newblock Robust unsupervised video anomaly detection by multipath frame prediction.
\newblock \emph{IEEE transactions on neural networks and learning systems}, 33\penalty0 (6):\penalty0 2301--2312, 2021.

\bibitem[Wu et~al.(2022)Wu, Hsieh, Chen, Fuh, and Liu]{wu2022self}
Jhih-Ciang Wu, He-Yen Hsieh, Ding-Jie Chen, Chiou-Shann Fuh, and Tyng-Luh Liu.
\newblock Self-supervised sparse representation for video anomaly detection.
\newblock In \emph{European Conference on Computer Vision}, pages 729--745. Springer, 2022.

\bibitem[Wu et~al.(2020)Wu, Liu, Shi, Sun, Shao, Wu, and Yang]{wu2020not}
Peng Wu, Jing Liu, Yujia Shi, Yujia Sun, Fangtao Shao, Zhaoyang Wu, and Zhiwei Yang.
\newblock Not only look, but also listen: Learning multimodal violence detection under weak supervision.
\newblock In \emph{European conference on computer vision}, pages 322--339. Springer, 2020.

\bibitem[Xu and Ding(2024)]{xu2024large}
Ruiyao Xu and Kaize Ding.
\newblock Large language models for anomaly and out-of-distribution detection: A survey.
\newblock \emph{arXiv preprint arXiv:2409.01980}, 2024.

\bibitem[Yang et~al.(2024)Yang, Lee, Dariush, Cao, and Lo]{yang2024anomalyruler}
Yuchen Yang, Kwonjoon Lee, Behzad Dariush, Yinzhi Cao, and Shao-Yuan Lo.
\newblock Follow the rules: Reasoning for video anomaly detection with large language models.
\newblock In \emph{Proceedings of the European Conference on Computer Vision (ECCV)}, 2024.

\bibitem[Yang et~al.(2025)Yang, Gao, Liu, Wu, Pang, and Shou]{yang2025assistpda}
Zhiwei Yang, Chen Gao, Jing Liu, Peng Wu, Guansong Pang, and Mike~Zheng Shou.
\newblock Assistpda: An online video surveillance assistant for video anomaly prediction, detection, and analysis.
\newblock \emph{arXiv preprint arXiv:2503.21904}, 2025.

\bibitem[Yao et~al.(2022)Yao, Wang, Xu, Pu, Wang, Atkins, and Crandall]{yao2022dota}
Yu Yao, Xizi Wang, Mingze Xu, Zelin Pu, Yuchen Wang, Ella Atkins, and David~J Crandall.
\newblock Dota: Unsupervised detection of traffic anomaly in driving videos.
\newblock \emph{IEEE transactions on pattern analysis and machine intelligence}, 45\penalty0 (1):\penalty0 444--459, 2022.

\bibitem[Ye et~al.(2025)Ye, Liu, and He]{ye2025vera}
Muchao Ye, Weiyang Liu, and Pan He.
\newblock Vera: Explainable video anomaly detection via verbalized learning of vision-language models.
\newblock In \emph{Proceedings of the Computer Vision and Pattern Recognition Conference}, pages 8679--8688, 2025.

\bibitem[Ye et~al.(2023)Ye, Xu, Xu, Ye, Yan, Zhou, Wang, Hu, Shi, Shi, et~al.]{ye2023mplug}
Qinghao Ye, Haiyang Xu, Guohai Xu, Jiabo Ye, Ming Yan, Yiyang Zhou, Junyang Wang, Anwen Hu, Pengcheng Shi, Yaya Shi, et~al.
\newblock mplug-owl: Modularization empowers large language models with multimodality.
\newblock \emph{arXiv preprint arXiv:2304.14178}, 2023.

\bibitem[Yuan et~al.(2023)Yuan, Zhang, Liu, Liu, Chen, Jin, and Jiao]{yuan2023surveillance}
Tongtong Yuan, Xuange Zhang, Kun Liu, Bo Liu, Chen Chen, Jian Jin, and Zhenzhen Jiao.
\newblock Towards surveillance video-and-language understanding: New dataset, baselines, and challenges, 2023.

\bibitem[Zaharescu and Wildes(2010)]{zaharescu2010anomalous}
Andrei Zaharescu and Richard Wildes.
\newblock Anomalous behaviour detection using spatiotemporal oriented energies, subset inclusion histogram comparison and event-driven processing.
\newblock In \emph{European Conference on Computer Vision}, pages 563--576. Springer, 2010.

\bibitem[Zanella et~al.(2024{\natexlab{a}})Zanella, Liberatori, Menapace, Poiesi, Wang, and Ricci]{zanella2024delving}
Luca Zanella, Benedetta Liberatori, Willi Menapace, Fabio Poiesi, Yiming Wang, and Elisa Ricci.
\newblock Delving into clip latent space for video anomaly recognition.
\newblock \emph{Computer Vision and Image Understanding}, 249:\penalty0 104163, 2024{\natexlab{a}}.

\bibitem[Zanella et~al.(2024{\natexlab{b}})Zanella, Menapace, Mancini, Wang, and Ricci]{zanella2024harnessing}
Luca Zanella, Willi Menapace, Massimiliano Mancini, Yiming Wang, and Elisa Ricci.
\newblock Harnessing large language models for training-free video anomaly detection.
\newblock In \emph{Proceedings of the IEEE/CVF Conference on Computer Vision and Pattern Recognition}, pages 18527--18536, 2024{\natexlab{b}}.

\bibitem[Zhang et~al.(2023{\natexlab{a}})Zhang, Li, and Bing]{damonlpsg2023videollama}
Hang Zhang, Xin Li, and Lidong Bing.
\newblock Video-llama: An instruction-tuned audio-visual language model for video understanding.
\newblock \emph{arXiv preprint arXiv:2306.02858}, 2023{\natexlab{a}}.

\bibitem[Zhang et~al.(2025)Zhang, Xu, Wang, Zuo, Huang, Gao, Zhang, Yu, and Sang]{zhang2025holmes}
Huaxin Zhang, Xiaohao Xu, Xiang Wang, Jialong Zuo, Xiaonan Huang, Changxin Gao, Shanjun Zhang, Li Yu, and Nong Sang.
\newblock Holmes-vau: Towards long-term video anomaly understanding at any granularity.
\newblock In \emph{Proceedings of the Computer Vision and Pattern Recognition Conference}, pages 13843--13853, 2025.

\bibitem[Zhang et~al.(2023{\natexlab{b}})Zhang, Han, Liu, Gao, Zhou, Hu, Yan, Lu, Li, and Qiao]{zhang2023llamaadapter}
Renrui Zhang, Jiaming Han, Chris Liu, Peng Gao, Aojun Zhou, Xiangfei Hu, Shilin Yan, Pan Lu, Hongsheng Li, and Yu Qiao.
\newblock Llama-adapter: Efficient fine-tuning of language models with zero-init attention.
\newblock \emph{arXiv preprint arXiv:2303.16199}, 2023{\natexlab{b}}.

\bibitem[Zhang et~al.(2024)Zhang, Wu, Li, Li, Ma, Liu, and Li]{zhang2024videoinstructiontuningsynthetic}
Yuanhan Zhang, Jinming Wu, Wei Li, Bo Li, Zejun Ma, Ziwei Liu, and Chunyuan Li.
\newblock Video instruction tuning with synthetic data, 2024.

\bibitem[Zhao et~al.(2019)Zhao, Peyrard, Liu, Gao, Meyer, and Eger]{zhao2019moverscore}
Wei Zhao, Maxime Peyrard, Fei Liu, Yang Gao, Christian~M Meyer, and Steffen Eger.
\newblock Moverscore: Text generation evaluating with contextualized embeddings and earth mover distance.
\newblock \emph{arXiv preprint arXiv:1909.02622}, 2019.

\bibitem[Zhao et~al.(2017)Zhao, Deng, Shen, Liu, Lu, and Hua]{zhao2017spatio}
Yiru Zhao, Bing Deng, Chen Shen, Yao Liu, Hongtao Lu, and Xian-Sheng Hua.
\newblock Spatio-temporal autoencoder for video anomaly detection.
\newblock In \emph{Proceedings of the 25th ACM international conference on Multimedia}, pages 1933--1941, 2017.

\bibitem[Zhou et~al.(2023)Zhou, Yu, and Yang]{zhou2023dual}
Hang Zhou, Junqing Yu, and Wei Yang.
\newblock Dual memory units with uncertainty regulation for weakly supervised video anomaly detection.
\newblock In \emph{Proceedings of the AAAI Conference on Artificial Intelligence}, pages 3769--3777, 2023.

\bibitem[Zhou et~al.(2024)Zhou, Qu, Xu, Shen, Song, and Shen]{zhou2024batchnorm}
Yixuan Zhou, Yi Qu, Xing Xu, Fumin Shen, Jingkuan Song, and Heng~Tao Shen.
\newblock Batchnorm-based weakly supervised video anomaly detection.
\newblock \emph{IEEE Transactions on Circuits and Systems for Video Technology}, 2024.

\bibitem[Zhu et~al.(2023)Zhu, Chen, Shen, Li, and Elhoseiny]{zhu2023minigpt}
Deyao Zhu, Jun Chen, Xiaoqian Shen, Xiang Li, and Mohamed Elhoseiny.
\newblock Minigpt-4: Enhancing vision-language understanding with advanced large language models.
\newblock \emph{arXiv preprint arXiv:2304.10592}, 2023.

\end{thebibliography}
}









\clearpage
\appendix

\section*{Supplementary Material}
\addcontentsline{toc}{section}{Supplementary Material}

\begingroup
  \renewcommand{\contentsname}{Contents}
  \setcounter{tocdepth}{2}

  \startcontents[supp]
  \printcontents[supp]{}{1}{\setcounter{tocdepth}{2}}
\endgroup
\bigskip

\section{Illustrative Example of Motivation}
\label{sec:supp_motivation}
\begin{figure*}[h!]
    \centering
    \includegraphics[width=\textwidth]{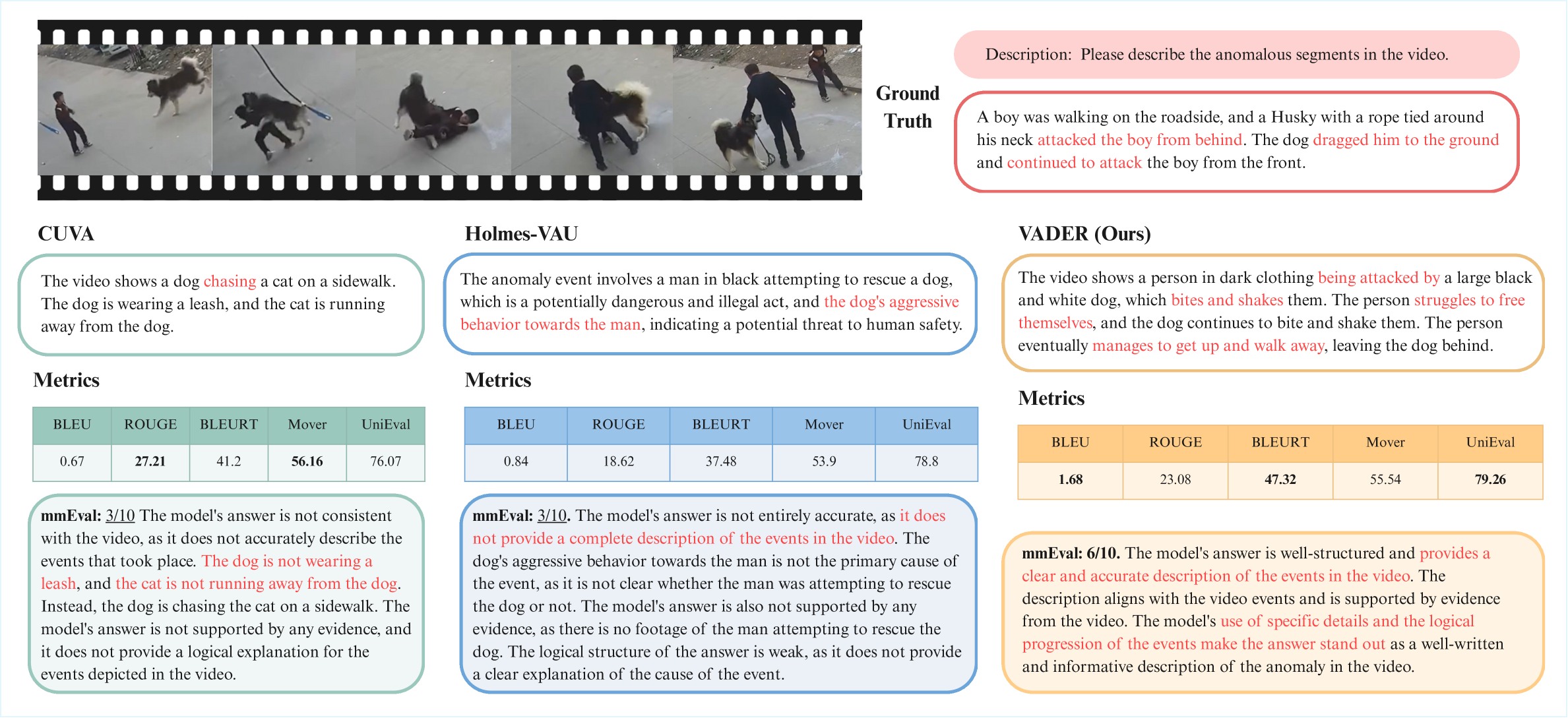}
    \caption{\textbf{Illustrative example showing the need for causal and relational modeling. } Given the same video of a dog attacking a boy, CUVA~\cite{CUVA} and Holmes-VAU~\cite{zhang2025holmes} produce incorrect or incomplete descriptions, while VADER captures key interactions and event progression, resulting in accurate and coherent descriptions with higher human-aligned evaluation scores.}
    \label{fig:motivation}
\end{figure*}

To better illustrate the motivation behind our work, Figure~\ref{fig:motivation} presents a real-world anomalous event where a dog suddenly attacks a boy walking on the roadside. This example demonstrates the challenges in understanding not only what happens in the scene, but also how it occurs, which requires modeling both causal relationships and dynamic object interactions.

In this case, different methods generate diverse outputs when asked to describe the anomalous segment:

\textbf{CUVA}~\cite{CUVA}
focuses on basic visual elements but produces a description that does not align with the actual event. While certain textual metrics such as ROUGE are relatively high, the description lacks factual accuracy and fails to capture the core abnormal interaction.

\textbf{Holmes-VAU}~\cite{zhang2025holmes} includes some relevant context, such as the presence of danger and a dog, but its narrative remains ambiguous. The description does not clearly reflect the cause-effect chain or the temporal progression of the event, limiting interpretability.

\textbf{VADER (Ours)} explicitly models the interactions and causal dynamics between objects, resulting in a coherent description that clearly states who was involved and how the anomaly unfolded. This produces outputs that align well with human judgments, as shown by higher semantic-level and human-evaluation metrics (e.g., BLEURT, UniEval, and mmEval).

The quantitative metrics and qualitative feedback below each output further highlight the importance of deeper reasoning. While lexical overlap metrics alone may not fully capture factual correctness, human-aligned evaluations reflect the clarity and interpretability of the descriptions. VADER achieves the highest ratings by providing a precise and logically structured explanation of the anomalous event.

\section{Ablation Study on Sample Mining Hyperparameters}
\label{sec:supp_ablation_hyp}
We perform an ablation study on the Gaussian smoothing parameter ($\sigma$) and the peak selection threshold (top-$k$ percentile) in our weakly-supervised sample mining pipeline. As shown in Table~\ref{tab:ablation_sigma_topk_auc}, we report AUC scores on the test set for different combinations of $\sigma$ and top-$k$.

Our results reveal two main findings. First, with $\sigma=2.0$, a top-$k$ percentile of $5\%$ yields the best AUC ($72.12$). Increasing the threshold to $7\%$ slightly reduces performance, likely due to the inclusion of less informative (noisy) samples. Conversely, a stricter threshold of $3\%$ also degrades performance, suggesting that overly selective sampling yields too few positives for effective training.
Second, across all thresholds, $\sigma=2.0$ consistently outperforms both smaller ($\sigma=1.0$) and larger ($\sigma=3.0$) values. We attribute this to improved noise filtering at $\sigma=2.0$ without excessive smoothing that would obscure salient events.

In summary, $\sigma=2.0$ and top-$k=5\%$ strike the best balance between sample quality and quantity, and are used in all subsequent experiments.

\begin{table}[ht]
\centering
\resizebox{0.7\columnwidth}{!}{
\begin{tabular}{@{}lccc@{}}
\toprule
\textbf{Smooth Sigma ($\sigma$)} & \multicolumn{3}{c}{\textbf{Top-k Percentile (\%)}} \\
\cmidrule(l){2-4}
 & \textbf{3.0\%} & \textbf{5.0\%} & \textbf{7.0\%} \\
\midrule
\textbf{1.0} & 69.53 & 70.18 & 68.91 \\
\textbf{2.0} & 70.85 & \textbf{72.12} & 71.64 \\
\textbf{3.0} & 68.17 & 69.25 & 68.55 \\
\bottomrule
\end{tabular}%
}
\caption{\textbf{Ablation study on sample mining hyperparameters.}
Test set AUC (\%) for different combinations of Gaussian smoothing parameter ($\sigma$) and top-$k$ percentile thresholds used in peak selection. $\sigma=2.0$ and top-$k=5\%$ achieve the best performance, highlighting the importance of balancing sample quality and quantity in our weakly-supervised mining strategy.}
\label{tab:ablation_sigma_topk_auc}
\end{table}

\section{Failure Cases}
\label{sec:supp_falure_case}

\begin{figure*}[h!]
    \centering
    \includegraphics[width=\textwidth]{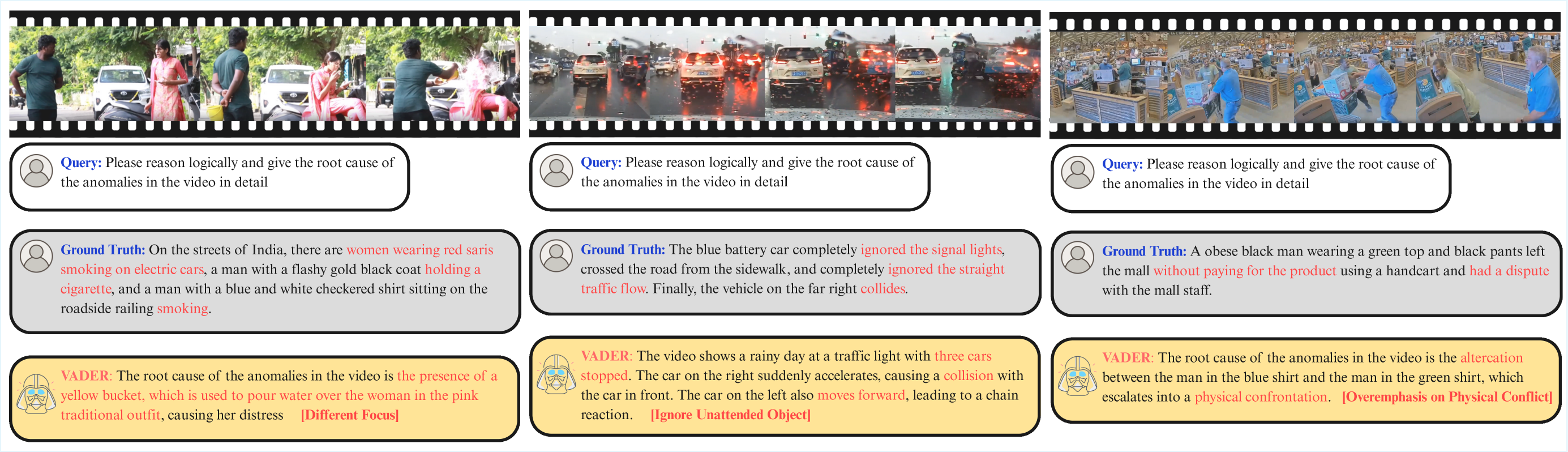}
    \caption{\textbf{Examples illustrating VADER's high-motion bias.} VADER tends to focus on visually prominent or dynamic actions, overlooking subtle or context-dependent cues. In the left example, it emphasizes pouring water while ignoring the public smoking. In the middle example, it identifies the car collision but misses the underlying cause, which is the vehicle running the traffic signal. In the right example, it overemphasizes the physical confrontation while neglecting the initial theft that triggered the anomaly.}
    \label{fig:motion_bias}
\end{figure*}

Figure~\ref{fig:motion_bias} presents three examples highlighting VADER’s tendency to favor visually dynamic events. While VADER accurately captures prominent high-motion activities, it often overlooks the underlying causes of anomalies or neglects subtle contextual cues, such as unattended objects or gradual environmental changes. For instance, in the middle example, VADER correctly identifies the car collision but misses the key causal factor, the vehicle ignoring the traffic signal, resulting in an incomplete explanation of why the anomaly occurred.

\begin{figure*}[h!]
    \centering
    \includegraphics[width=\textwidth]{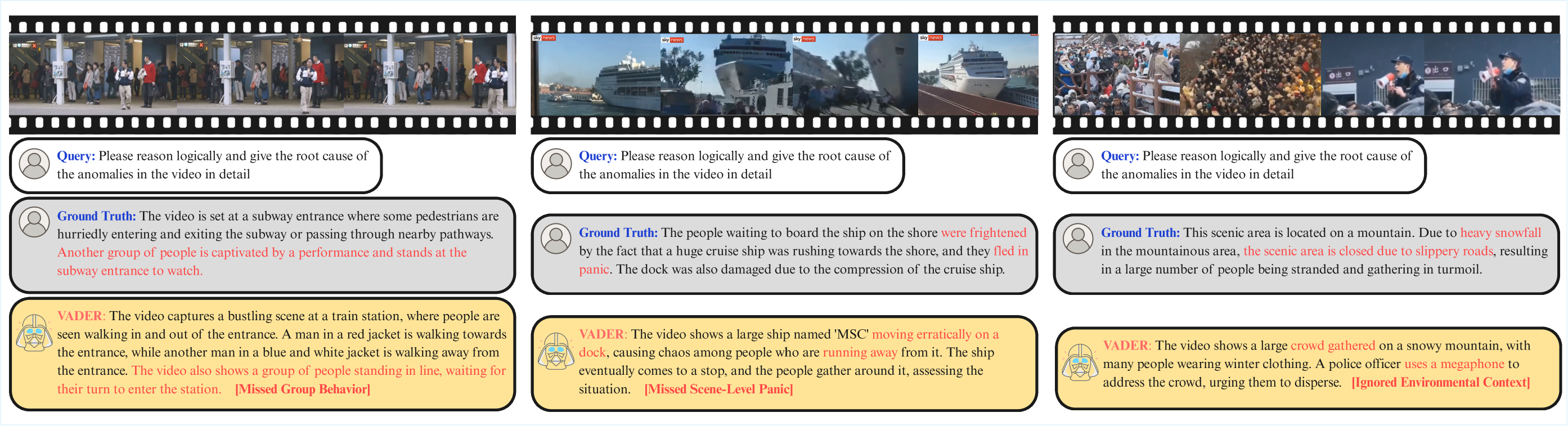}
    \caption{\textbf{Examples illustrating VADER's limitation to object-centric reasoning.} VADER focuses on localized pairwise interactions, overlooking group behaviors or scene-level environmental factors. In the left example, it describes individual pedestrian movements but misses the crowd gathered to watch a performance. In the middle example, it identifies the ship's movement but overlooks the collective panic it causes. In the right example, it focuses on the police officer and the crowd but ignores the environmental cause behind the stranded people.}
    \label{fig:object_centric}
\end{figure*}

Figure~\ref{fig:object_centric} presents three examples highlighting VADER’s limitation to object-centric reasoning. While VADER accurately captures localized actions and pairwise object interactions, it often fails to represent the broader scene context or collective behaviors. For instance, in the left example, VADER correctly describes individual pedestrians entering and exiting the subway but misses the group of people gathered to watch a street performance, resulting in an incomplete understanding of the true anomaly. 

\section{Potential Mitigations for Limitations}
\label{sec:supp_mitigation_for_limitations}

To address the three primary limitations discussed in Sec. 5, we outline potential mitigation strategies. These strategies aim to enhance VADER’s robustness, reduce bias, and broaden its capacity to handle diverse anomaly scenarios.

\subsection{Mitigating Dependency on Upstream Modules}
VADER relies on the performance of upstream modules, which may cause errors to propagate and impact the final reasoning results. To reduce this dependency and improve the reliability of the pipeline, we consider the following approaches to strengthen upstream components and improve their alignment with downstream tasks:

\textbf{End-to-End Joint Training.}
Transform the current pipeline into a jointly trainable framework, allowing gradients from the LLM to update the upstream modules so they can better align with the final reasoning objectives.

\textbf{Module Robustness Enhancement.}
Before end-to-end integration, each module can be strengthened individually by training on more diverse data and employing more advanced tracking algorithms, thereby reducing upstream errors and improving overall robustness

\subsection{Reducing Bias Toward High-Motion Events}
VADER’s reliance on relational volatility can lead to a bias toward anomalies involving strong motion while overlooking subtle anomalies. To address this imbalance, additional cues can be incorporated to complement volatility:

\textbf{Object State Modeling.}
Track and interpret object states, such as transitions from “carried” to “stationary” without an owner, to capture static anomalies like abandoned objects.

\textbf{Global Scene Context.}
Model typical activity flow within the scene and identify deviations, enabling the detection of subtle anomalies such as loitering or suspicious inactivity.

\subsection{Moving Beyond Object-Centric Reasoning}
The current design of VADER focuses on pairwise object interactions, limiting its ability to capture scene-level or group-level anomalies.
To broaden the scope of anomaly understanding, the following extensions can be considered:

\textbf{Global Scene Modeling.}
Add a global feature stream that directly encodes entire frames to detect anomalies not tied to specific objects, such as lighting changes or smoke. 

\textbf{Group-Level Reasoning.}
Identify and model groups of objects to capture collective behaviors, allowing detection of emergent events like crowd panic or mass movements.

\section{Computational Efficiency}
\label{sec:supp_computational_efficiency}

We provide a comparison of inference speed between VADER and NVILA~\cite{liu2025nvila} on the HAWK benchmark in Table~\ref{tab:efficiency}. The table reports both the total inference time (in minutes) and frames per second (fps). Although VADER introduces additional modules for relational reasoning and context-aware sampling, it still operates at a practical speed for real-world applications.

\begin{table}[h]
\centering
\resizebox{0.7\columnwidth}{!}{
\begin{tabular}{lcc}
\toprule
\textbf{Method} & \textbf{Total Time (min)} & \textbf{FPS} \\
\midrule
NVILA~\cite{liu2025nvila} & 49.85 & 33.63 \\
VADER & 87.25 & 19.22 \\
\bottomrule
\end{tabular}
}
\caption{Comparison of inference time and fps between NVILA~\cite{liu2025nvila} and VADER on the HAWK~\cite{atang2024hawk} benchmark.}
\label{tab:efficiency}
\end{table}

\section{Intermediate Visualization of Module Contributions}
\label{sec:supp_intermediate_vis}

\begin{figure}[h]
    \centering
    \includegraphics[width=\linewidth]{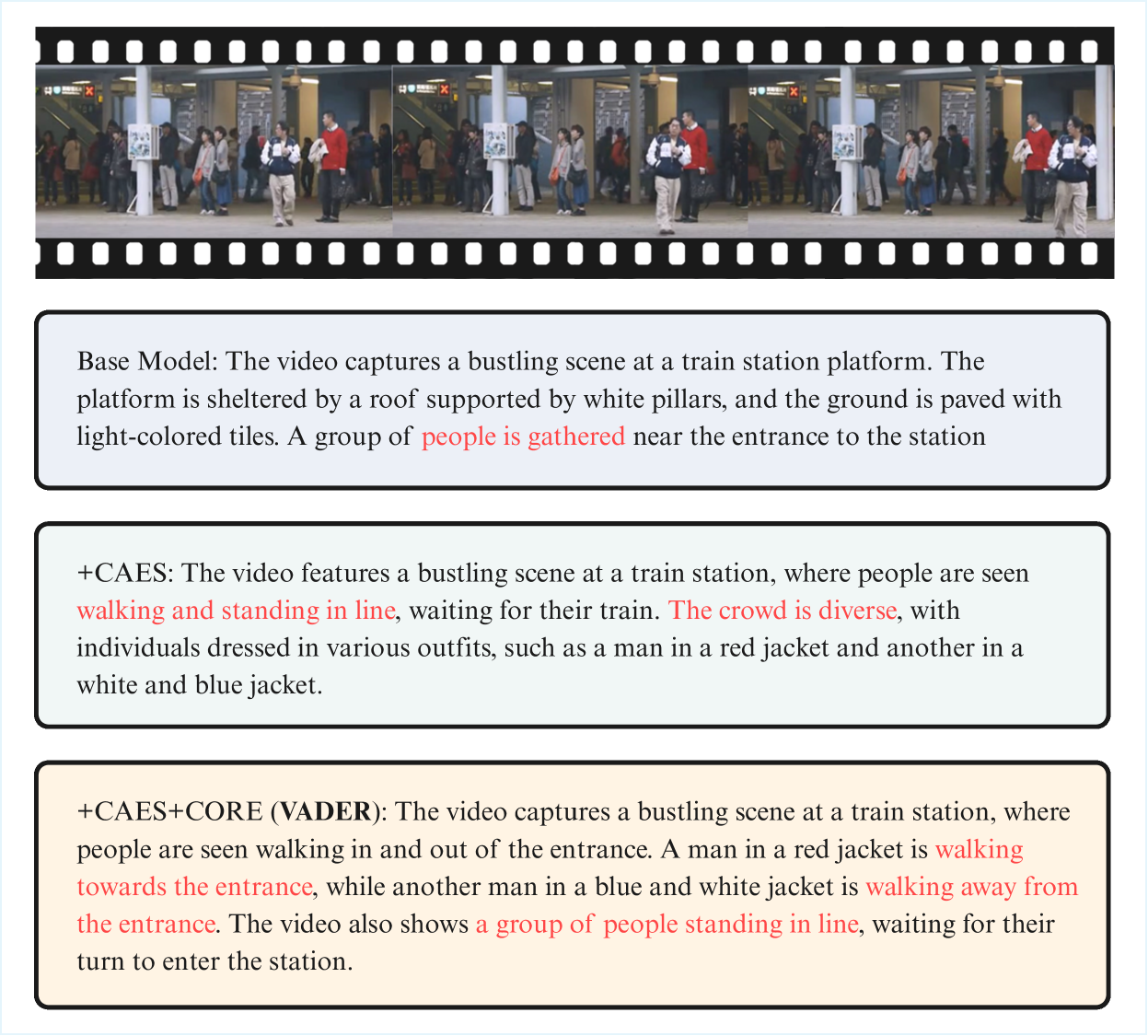}
    \caption{\textbf{Intermediate visualization} showing how CAES and CORE improve descriptions. CAES helps focus on key frames, while CORE models object interactions for more interpretable outputs.}
    \label{fig:Intermediate-visualization}
\end{figure}

Figure~\ref{fig:Intermediate-visualization} illustrates how each module incrementally improves the generated descriptions. The example shows a bustling scene at a train station. The Base Model generates a generic narrative without focusing on key details. CAES helps the model attend to important frames, capturing activities like walking and standing in line. CORE further models object interactions, enabling it to distinguish directional actions and explain relationships between people and their surroundings. This step-by-step progression demonstrates how each module contributes to improving both descriptive accuracy and interpretability.

\section{Implementation Details}
\label{sec:supp_implementaiton_details}

Anomaly intervals are detected using an adaptive threshold with the 97th percentile computed per video. For each interval, pre- and post-event contexts are determined by the 95th and 85th percentiles of the score slope over a 5-frame window, with a maximum context window of 30 frames. We uniformly sample 4 frames from each context, 8 from the event, and fill to 64 frames with background frames.

For relational analysis, object association combines cosine similarity of appearance embeddings (weight 0.8) and IoU, with matching solved by the Hungarian algorithm~\cite{kuhn1955hungarian} and a maximum track age of 15 frames. Relational volatility at each timestep is measured as the maximum L2 distance between all co-tracked relation pairs in adjacent frames, followed by Gaussian smoothing with a standard deviation of 2.0. The top 5\% of volatility peaks are used as positives.

The Relational Dynamic Encoder is a two-layer MLP trained with triplet margin loss~\cite{schroff2015facenet} with margin 0.5 and semi-hard negative mining~\cite{schroff2015facenet} with pool size 30. The encoder training uses Adam (learning rate $1\times10^{-4}$) with StepLR for 50 epochs.

For LLM fine-tuning, we adopt NVILA~\cite{li2025otter} as the backbone, updating only the projector and LoRA~\cite{hu2022lora} adapters while freezing all other parameters. The learning rate is set to $2\times10^{-5}$ with a cosine schedule and a warm-up ratio of 0.03.



\end{document}